\documentclass[lettersize,journal]{IEEEtran}
\usepackage{amsmath,amsfonts}
\usepackage{algorithmic}
\usepackage{array}

\usepackage[caption=false,font=normalsize,labelfont=rm,textfont=rm]{subfig}

\usepackage{textcomp}
\usepackage{stfloats}
\usepackage{url}
\usepackage{verbatim}
\usepackage{graphicx}
\def\BibTeX{{\rm B\kern-.05em{\sc i\kern-.025em b}\kern-.08em
    T\kern-.1667em\lower.7ex\hbox{E}\kern-.125emX}}
\usepackage{balance}

\usepackage{colortbl}
\usepackage{xcolor}
\definecolor{mygray}{gray}{.9}
\usepackage{multirow}
\usepackage{tablefootnote}
\definecolor{revised}{RGB}{255, 0, 0}
\definecolor{revised_checked}{RGB}{0, 0, 0}
\definecolor{revised_r1}{RGB}{0, 0, 0}

\usepackage{enumitem}

\usepackage[pagebackref=true,breaklinks=true,letterpaper=true,colorlinks,bookmarks=false]{hyperref}

\usepackage{amssymb}

\begin{document}
\title{\textcolor{revised_checked}{SETA: Semantic-Aware Edge-Guided Token Augmentation for Domain Generalization}}
\author{Jintao Guo, Lei Qi, Yinghuan Shi*, Yang Gao
\thanks{
Jintao Guo, Yinghuan Shi, and Yang Gao are with the National Key Laboratory for Novel Software Technology and the National Institute of Healthcare Data Science, Nanjing University, Nanjing, China,
210023 (e-mail: guojintao@smail.nju.edu.cn; syh@nju.edu.cn; gaoy@nju.edu.cn).

Lei Qi is with the School of Computer Science and Engineering, Southeast University, and Key Laboratory of New Generation Artificial Intelligence Technology and Its Interdisciplinary Applications (Southeast University), Ministry of Education, Nanjing, China, 211189 (e-mail: qilei@seu.edu.cn).

*The corresponding author: Yinghuan Shi.

This work was supported by the National Key R\&D Program of China (2023ZD0120700, 2023ZD0120701), NSFC Project (62222604, 62206052, 62192783), 
China Postdoctoral Science Foundation (2024M750424), the Fundamental Research Funds for the Central Universities (020214380120), the State Key Laboratory Fund (ZZKT2024A14), the Postdoctoral Fellowship Program of CPSF (GZC20240252), and the Jiangsu Funding Program for Excellent Postdoctoral Talent (2024ZB242).
}}



\maketitle

\begin{abstract}
    Domain generalization (DG) aims to enhance the model robustness against domain shifts without accessing target domains. A prevalent category of methods for DG is data augmentation, which focuses on generating virtual samples to simulate domain shifts. However, existing augmentation techniques in DG are mainly tailored for convolutional neural networks (CNNs), with limited exploration in token-based architectures, \textit{i.e.}, vision transformer (ViT) and multi-layer perceptrons (MLP) models. In this paper, we study the impact of prior CNN-based augmentation methods on token-based models, revealing their performance is suboptimal due to the lack of incentivizing the model to learn holistic shape information. \textcolor{revised_checked}{To tackle the issue, we propose the Semantic-aware Edge-guided Token Augmentation (SETA) method. SETA transforms token features by perturbing local edge cues while preserving global shape features, thereby enhancing the model learning of shape information.} To further enhance the generalization ability of the model, we introduce two stylized variants of our method combined with two state-of-the-art (SOTA) style augmentation methods in DG. We provide a theoretical insight into our method, demonstrating its effectiveness in reducing the generalization risk bound. Comprehensive experiments on five benchmarks prove that our method achieves SOTA performances across various ViT and MLP architectures. Our code is available at \textcolor{magenta}{\href{https://github.com/lingeringlight/SETA}{https://github.com/lingeringlight/SETA}}. 
\end{abstract}



\section{Introduction}
\IEEEPARstart{I}{n} recent years, deep neural networks have achieved remarkable success in various vision tasks \cite{kirillov2023segment,zhao2024learning,yan2023clip}. 
These models mainly rely on the independent and identically distributed (\textit{i.i.d.}) assumption that training and test data are sampled from the same distribution \cite{pan2009survey}.
However, in real-world scenarios, the \textit{i.i.d.} assumption does not always hold, leading to severe degradation in model performance when dealing with test data that follows different distributions from the training data (\textit{a.k.a}, domain shift \cite{ben2010theory}).
To tackle the issue, domain generalization (DG) \cite{zhou2022domain} has  
emerged as a feasible solution, which aims to train a model from multiple observed source domains that generalizes well to unseen target domains without re-training. 
Existing DG methods primarily utilize domain alignment \cite{gholami2023latent,guo2023domaindrop}, meta-learning \cite{wang2023generalizable,zhang2023style}, and data augmentation \cite{li2023cross,wang2024inter}, which have shown promising results.

\begin{figure}[tb!]
  \begin{center}
  \includegraphics[width=\linewidth]{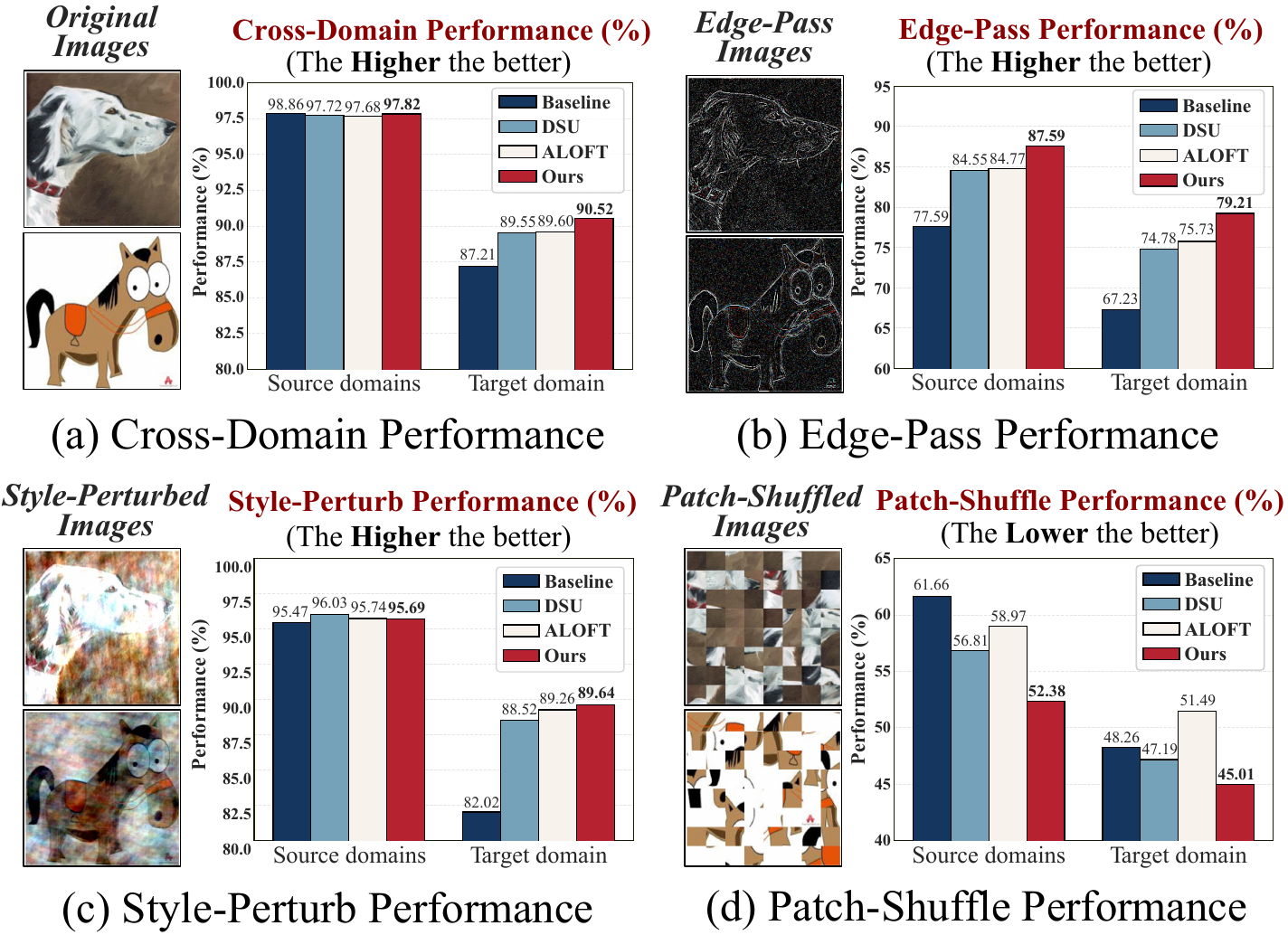}
  \end{center}
  \vspace{-0.2cm}
  \caption{\textcolor{revised_checked}{Comparisons of models on cross-domain performance, edge bias, shape bias, and style shift robustness. 
  We compare our method with SOTA augmentation methods in DG (ALOFT \cite{guo2023aloft} and DSU \cite{li2022uncertainty}) on 
  (a) original samples; (b) edge-pass samples reconstructed from phase spectrums \cite{bai2022improving}; (c) style-perturbed samples with amplitude spectrums perturbed \cite{fridovich2022spectral}; (d) patch-shuffled samples with shuffled patch locations within sample \cite{baker2018deep}. 
  Experiments are conducted on PACS with the backbone GFNet-H-Ti \cite{rao2021global}.}
  \label{fig:bias acc}}
  \vspace{-0.5cm}
\end{figure}

Among recent DG methods, data augmentation methods have exhibited promising performance \cite{zhou2023mixstyle,li2023cross,guo2023aloft,li2022uncertainty}. 
These methods primarily assume that domain gaps are caused by the disparity in style, thus introducing channel-level style perturbations for simulating potential domain shifts.
While these methods have achieved notable success on CNN architectures, limited research investigates their effects on ViT or MLP models. 
\textcolor{revised_checked}{
ViT and MLP models have shown excellent generalization ability \cite{guo2023aloft,ren2023masked}, 
attributed to their capacity to learn spatial dependencies among tokens, which encode the structural shape information of objects, thus exhibiting a superior shape bias than CNNs \cite{guo2023aloft,bai2022improving}. 
 Current DG augmentation methods focus on perturbing style information, which has proven particularly effective for CNNs due to their severe texture bias.
 However, for token-based models (ViTs and MLPs) with inherently lower texture bias, 
 it remains uncertain \textit{whether augmenting style information isolatedly could sufficiently enhance their generalization.}
 Besides, previous methods mainly generate perturbations at the channel level, whereas token-based models focus on learning dependencies among tokens at the spatial level.
 Due to the lack of token-level augmentation for DG, the model could inevitably learn and extract domain-specific information from class-irrelevant tokens.
 Therefore, how to design an effective token-level augmentation for DG to enhance the generalization of token-based models sufficiently remains an open question. 
}



In this paper, we initially analyze the impact of existing SOTA augmentation methods in DG on the shape bias of the models. 
Specifically, we evaluate the robustness of these methods to diverse biased perturbations, focusing on two representative advanced methods: DSU \cite{li2022uncertainty}, which is based on statistical perturbation, and ALOFT \cite{guo2023aloft}, which exploits frequency space augmentation. 
The experiments are conducted on the representative MLP-like model GFNet \cite{rao2021global}.
As shown in Fig.~\ref{fig:bias acc} (a), both DSU and ALOFT methods exhibit the ability to enhance the model generalization.
The enhancement is attributed to their perturbations on style features, implicitly boosting the ability of the model to capture edge information, as indicated in Fig.~\ref{fig:bias acc} (b).
Moreover, the enhanced edge-capturing ability also contributes to the increased model robustness against style perturbations, as depicted in Fig.~\ref{fig:bias acc} (c). 
\textcolor{revised_checked}{
However, as identified in \cite{tripathi2023edges,hermann2020shapes}, \textit{image edges encompass not only global object shapes (\textit{i.e.}, category-related semantics), but also local spurious edges (\textit{i.e.}, category-irrelated edges like backbone textures).}
Since existing style augmentation methods for DG lack spatial token perturbation, they fail to suppress the domain-specific local edges, resulting in the limited ability of the model to perceive the holistic shape structures of objects, as proved in Fig.~\ref{fig:bias acc} (d).
Consequently, the model inevitably overfits the tokens encoding domain-specific features, which severely impedes its generalizability to unseen domains. }
These observations prompt a critical question:
\textit{Could we augment edge features at the token level to explicitly enhance the shape bias of the model,} thus further improving the generalization ability of ViT and MLP architectures?

To this end, we propose a novel Semantic-aware Edge-guided Token Augmentation (SETA) for DG, aiming at enhancing the generalization performance of ViT and MLP models by introducing token-level local edge perturbations.
Unlike previous DG augmentation methods that focus on diversifying channel-level styles, 
\textit{our method concentrates on generating token-level edge perturbations to guide the model in emphasizing global shape features during training.}
\textcolor{revised_checked}{
Specifically, we initially design the \textit{Activation-based Edge Tokens Selection} (ETS) module, which extracts edge maps from input features using low-pass filters and calculates the activation associated with each token to distinguish the edge-relevant tokens.
Subsequently, random pairs of samples are selected from the batch, 
with one sample retaining only the edge tokens and the other being modified by the \textit{Shape Tokens Shuffling} (STS) module that shuffles token locations within the sample.}
Finally, we conduct the random mixing of the two samples using either Mixup or CutMix, with the augmented sample being assigned the label of the sample providing edge tokens.
Furthermore, we extend our method by integrating SOTA style augmentation techniques to enhance both shape sensitivity and robustness to style shifts.
We evaluate our methods on five challenging datasets, employing various ViT and MLP architectures.
The results prove the effectiveness of our method to enhance model generalization.
Our contributions are summarized as follows:
\begin{itemize}[itemsep=3pt,topsep=3pt]
    \item We propose a token-level augmentation method, namely SETA, that explicitly incentivizes shape bias for domain generalization, which can enhance the generalization ability of various ViT and MLP architectures.
    \item We extend the SETA method with SOTA style-augmented DG methods to two stylized variants, proving that shape augmentation and style diversification can synergistically boost the model robustness to domain shifts.
    \item We theoretically prove that incentivizing shape sensitivity through our SETA could tighten the generalization risk bound and improve the generalizability of the model.
\end{itemize}

\section{Related Work}
\label{sec:Related Work}
\textbf{Domain generalization.} 
Domain generalization (DG) aims to learn generalized representations from source domains that generalize well to arbitrary unseen target domains. 
Existing DG methods address domain shift from various perspectives, including domain alignment \cite{wang2023sharpness,hemati2023understanding}, meta-learning \cite{wang2023generalizable}, data augmentation \cite{qi2024normaug}, ensemble learning \cite{zhang2022mvdg,arpit2021ensemble} and regularization methods \cite{zhang2023flatness,guo2023place,zhang2024exploring}.
The majority of relevant works to our method address the DG issue by data augmentation.
These methods operate under the assumption that domain gaps predominantly arise from style shifts, thereby endeavoring to simulate domain shifts by generating diverse styles while keeping semantics consistency.
Specifically, one prevalent line is to perturb the feature statistics (\textit{i.e.}, mean and variance) by sample interpolation \cite{zhou2023mixstyle,zhang2022exact}, modeling distribution \cite{li2022uncertainty} or learning methods \cite{kang2022style,fu2023styleadv}. 
Moreover, some recent works propose to leverage the semantic-preserving property of the Fourier phase component and introduce diverse noise into the amplitude spectrum to generate stylized data \cite{xu2021fourier,wang2022domain}.
\textcolor{revised_checked}{
However, these methods are mainly designed for CNN architectures, which focus on perturbing channel-level style features. 
When applied to ViT or MLP models that learn spatial dependencies among different tokens, 
as these methods lack token-level augmentation, the model inevitably overfits domain-specific tokens that encode spurious local edge information, which impedes its generalizability. 
Although some pioneering works \cite{sultana2022self,zheng2022prompt} have recently explored ViT-based methods for DG, these methods rely on supervisory signals from additional tokens to implicitly constrain the model learning of domain-invariant features, ignoring the presence of domain-related information within feature tokens. 
Due to the lack of explicit constraints on feature tokens, when facing significant domain shifts, these tokens are likely to contain substantial domain-specific information, leading to overfitting to source domains.
Therefore, we propose a token-level augmentation method that explicitly enhances domain-invariant feature extraction within feature tokens, which is the first DG method to improve the generalization ability of the model from the perspective of incentivizing global shape bias. 
}

\begin{figure*}[tb!]
    \begin{center}
    \includegraphics[width=\linewidth]{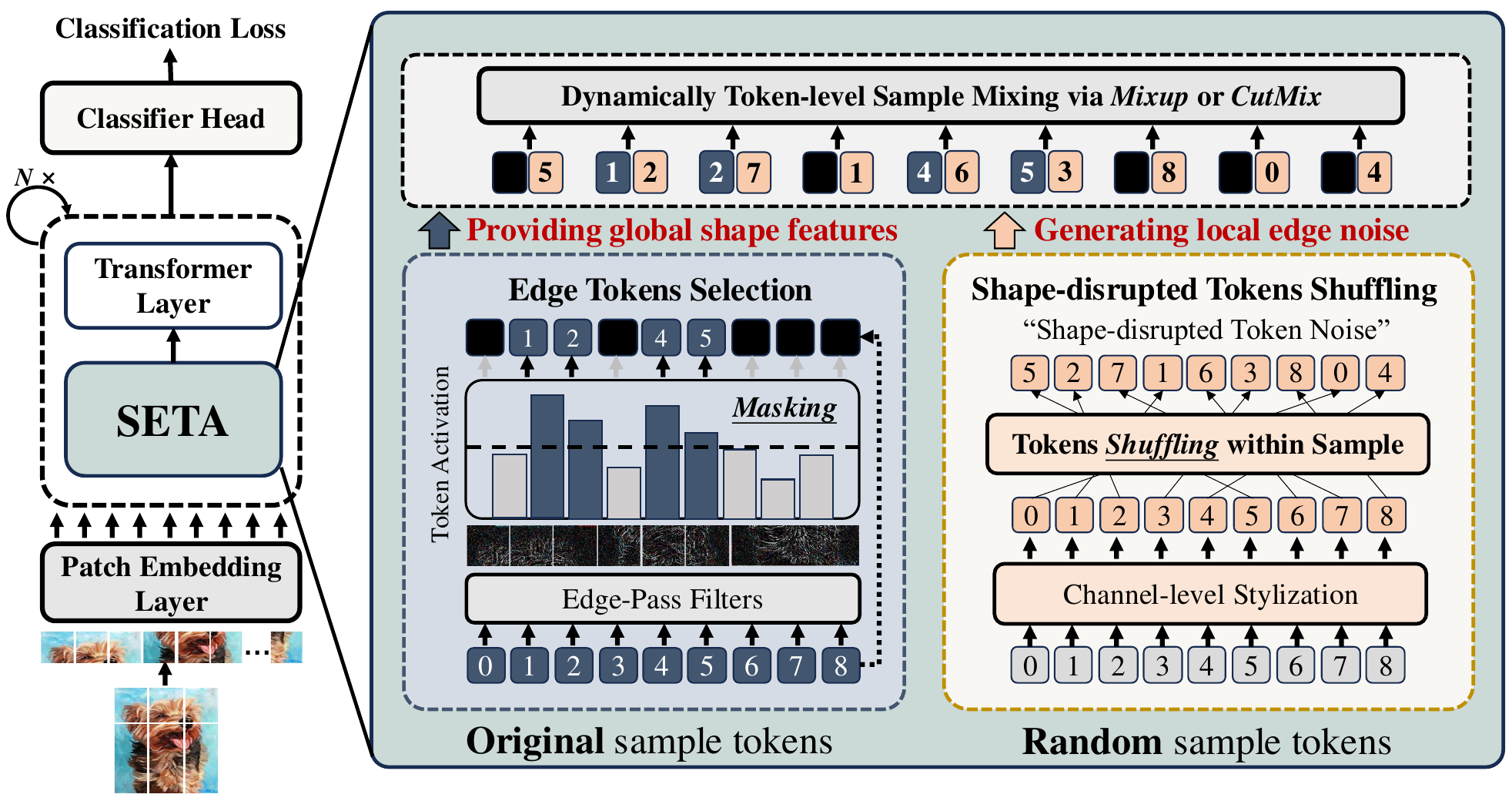}
    \end{center}
    \vspace{-0.2cm}
    \caption{
        \textcolor{revised_checked}{
        An overview of the proposed SETA. Our SETA consists of three core modules, including 1) Activation-based Edge Tokens Selection (ETS) that distinguishes and extracts tokens containing edge information; 2) Shape Tokens Shuffling (STS) that generates texture noise by shuffling tokens from another sample, which disrupts holistic shape while keeping local edges; 3) Token Mixing module that superposes edge tokens from an object sample onto the shuffled tokens from another sample. The augmented sample is assigned the label of the sample contributing the edge tokens. 
        We design two stylized variants of SETA, utilizing SOTA DG augmentation methods, \textit{i.e.}, DSU \cite{li2022uncertainty} and ALOFT \cite{guo2023aloft}, to create stylized token-shuffled samples to simulate potential domain shifts.}
    }
    \label{fig:framework}
    \vspace{-0.2cm}
\end{figure*}

\vspace{0.1cm}
\noindent\textbf{Data augmentation for improving shape bias.} 
Previous studies have revealed that compared to human vision, CNNs are revealed to be more sensitive to image texture for object classification, indicating that the trained models lack a full understanding of different object categories \cite{hermann2020shapes,geirhos2018imagenet}.
To reduce this disparity, augmentation methods aim to perturb texture representations, which could explicitly reduce the texture bias and implicitly boost the shape bias of the model \cite{hermann2020shapes}. 
\textcolor{revised_checked}{
Several works also indicate that an increased shape bias in CNNs is associated with improved model robustness \cite{shi2020informative,li2020shape}.
The exploration of texture and shape biases has extended to ViTs and MLPs, showing that they exhibit a lower texture bias compared to CNNs \cite{bai2022improving}. 
It is attributed to the ability of token-based models to effectively learn dependencies among tokens, thereby enhancing their capability to capture global semantic information. 
Some advanced token-level augmentations have been explored to improve the learning capacity of ViTs. 
TokenMixup \cite{choi2022tokenmixup} utilizes attention maps within ViTs to construct a saliency-aware token mixup for supervised classification, aiming to help the model learn label-related class information from tokens. 
PMTrans \cite{zhu2023patch} employs a game-theory-based learnable beta distribution to sample and mix patches from source and target domains, aligning the labeled source domain with the unlabeled target domain.
However, in DG scenarios, these methods cannot guarantee the model's performance on unseen target domains. 
Due to the lack of explicit constraints on domain-specific texture information, these methods inevitably suffer from texture bias, significantly impairing the model's ability to generalize across different domains. 
Although some DG methods \cite{zhou2023mixstyle,xu2021fourier} attempt to perturb style information while keeping the others unchanged (\textit{e.g.}, edges), they fail to suppress spurious edge information (\textit{i.e.}, shape-irrelevant textures), limiting their ability to enhance the shape bias.
In the paper, we propose a semantic-aware edge-guided token augmentation method, which can discern and perturb spurious edges while retaining crucial shape information. 
Our method compels the model to learn a comprehensive understanding of object shape from the edge-perturbed representations, thereby enhancing its shape bias effectively.}

\section{Method}

\subsection{Setting and Overview}
Given a training set of $K$ observed source domains $\mathcal{D}_s = \{D_s^1, D_s^2, ..., D_s^K\}$ with $n_k$ labeled samples $\{(x_i^k, y_i^k)\}_{i=1}^{n_k}$ in the $k$-th domain $D_s^k$ ($1 \leq k \leq K$), where $(x_i^k, y_i^k)$ is the sample-label pair for the $i$-th sample in the $k$-th domain. 
The goal of DG is to learn a domain-agnostic model on source domains $\mathcal{D}_s$ that generalizes well to arbitrary unseen target domain $\mathcal{D}_t$ without predefined distribution.

In standard DG model training, the model is directly trained to predict the label of the original samples. 
However, as emphasized in \cite{geirhos2018imagenet,li2020shape,islam2021shape}, this training method could induce sensitivity to style and texture in source domains, consequently hindering model generalization on unseen target domains. 
\textcolor{revised_checked}{
To tackle this issue, we propose a novel token-level augmentation method, namely Semantic-aware Edge-guided Token Augmentation (SETA), for ViT and MLP models. 
The key idea of SETA is to generate shape-enhanced representations by mixing edge-related tokens from a sample and shape-disrupted tokens from other samples.
This operation compels the model to distinguish the overall shape of objects from the shape-irrelated edge-perturbed representations, thereby enhancing its sensitivity to shape information.}
To further enhance the model robustness to domain shifts, we explore a paradigm that integrates SETA with SOTA augmentation methods in DG, yielding two stylized variants of SETA.
An overview of our method is depicted in Fig.~\ref{fig:framework}.
Below we introduce each component of our method and provide a theoretical insight into its effectiveness for improving model generalization.

\subsection{\textcolor{revised_checked}{Activation-based Edge Tokens Selection}} 

\textcolor{revised_checked}{
Previous works \cite{tripathi2023edges,hermann2020shapes} have revealed that image edges encompass both global shape structures (\textit{i.e.}, category-related object shapes) and local edge textures (\textit{i.e.}, category-irrelevant local edges). 
However, directly distinguishing shape concepts from images is challenging. 
To address this, our method focuses on \textit{perturbing local edge textures to implicitly enhance the model's ability to learn the remaining shape structure information.} 
Specifically, following \cite{guo2023aloft,wang2022domain}, we first design the \textit{Activation-based Edge Token Selection (ETS)} module to extract edge information, \textit{i.e.}, leveraging Fourier transformation to extract high-frequency component of representation.
}

\vspace{0.1cm}
\textbf{Extracting edge representations.} Given the input feature map $z_i \in \mathbb{R}^{N \times C}$, where $N$ is the number of tokens and $C$ is the number of channels. 
We first reshape the input feature into $z_i \in \mathbb{R}^{H \times W \times C}$, where $H$ and $W$ are equal to $\lfloor \sqrt{N}\rfloor $. Then, we obtain its Fourier transformation:
\begin{equation}
    \mathcal{F}(z_i)_{u, v, c} = \sum_{h=0}^{H-1} \sum_{w=0}^{W-1} z_i(h, w, c) e^{-j2\pi(\frac{h}{H}u + \frac{w}{W}v)}.
    \label{eq:FFT}
\end{equation}

The frequency signal $\mathcal{F}(z_i)$ is then decomposed into the amplitude component $\mathcal{A}(z_i) \in \mathbb{R}^{H \times W \times C}$ and the phase component $\mathcal{P}(z_i)\in \mathbb{R}^{H \times W \times C}$, where the amplitude spectrum contains low-level statistics (\textit{e.g.}, style information) while the phase spectrum encodes high-level semantics (\textit{e.g.}, edge and shape). 
With the low-frequency components being shifted to the center of the amplitude spectrum, we employ a binary mask $\mathcal{M}_r \in \mathbb{B}^{H \times W}$, whose value is zero only for the center region of scale $r$, to control the scale of low-frequency component to be discarded. 
Afterwards, we can obtain the high-frequency amplitude spectrum $\hat{\mathcal{A}}(z_i)$:
\begin{equation}
    \hat{\mathcal{A}}(z_i)_{u, v, c} = \mathcal{M}_r \odot \mathcal{A}(z_i)_{u, v, c},
\end{equation}
where $\odot$ is element-wise multiplication. 
The amplitude spectrum $\hat{\mathcal{A}}(z_i)$ is combined with the original phase spectrum $\mathcal{P}(z_i)$ to form the edge-pass Fourier representation:
\begin{equation}
    \hat{\mathcal{F}}(z_i)_{u, v, c} = \hat{\mathcal{A}}(z_i)_{u, v, c} * e^{-j*\mathcal{P}(z_i)_{u, v, c}}.
\end{equation}
The inverse Fourier transformation $\mathcal{F}^{-1}$ is then applied to generate the edge-pass representation: $\hat{z}_i = \mathcal{F}^{-1}[\hat{\mathcal{F}}(z_i)]$, which is reshaped back to $\hat{z}_i \in \mathbb{R}^{N \times C}$.

\vspace{0.1cm}
\textcolor{revised_checked}{
\textbf{Selecting edge tokens by activation.} Since $\hat{z}_i$ primarily retains edge-related information, to quantify the extent of edge information contained within each token, we calculate its activation value within $\hat{z}_i$, which is formulated as:
\begin{equation}
    A(\hat{z}_i)_{\text{token}_j} = \frac{1}{C} \sum_{c=0}^{C-1} \hat{z}_i(j, c).
\end{equation}
The larger the activation value, the more edge information the token contains.
With the token-level activation $A(\hat{z}_i)_{\text{token}_j}$, we compute the $p$-th percentile, denoted as $q_{p}$, and perform a thresholding operation to obtain the edge-information guided mask $\mathcal{M}_{\text{edge}}\in \mathbb{B}^{N \times C}$. The element is set to $1$ if the corresponding element in $A(\hat{z_i})$ falls within the top $p$ percentage elements of $A(\hat{z}_i)$, and set to $0$ otherwise: $\mathcal{M}_{\text{edge}} (j, c) = \mathbb{I} (A(\hat{z}_i)_{\text{token}_j} \geq q_{p}).$}
Finally, we could distinguish and extract the edge tokens from the input features $z_i$ with the binary mask $\mathcal{M}_{\text{edge}}$, which is formulated as:
\begin{equation}
    z^{e}_i = \mathcal{M}_{\text{edge}} \odot z_i.
    \label{Eq:edge tokens}
\end{equation}

Noting that in the process, the low-frequency filtered representation $\hat{z}_i$ is leveraged to generate the token-level mask $\mathcal{M}_{\text{edge}}$, which is applied to the original representation $z_i$ for preserving global shape features. 
The motivation behind this operation is that the low-frequency filtering could also lead to the loss of crucial semantic information surrounding edges, which may impede model training.
Thanks to our proposed activation-based edge tokens selection (ETS), the augmented representations could retain critical object information within the edge-related tokens while discarding the edge-irrelevant tokens.
We also evaluate the strategy that directly utilizes low-frequency filtering representation $\hat{z}_i$ to provide shape features in Sec.~\ref{exp:EST strategy}. The results validate the effectiveness of our ETS in preserving useful semantic information around object edges while suppressing model learning of domain-specific features.


\subsection{Enhancing Shape Bias by Texture Noise} 

\textcolor{revised_checked}{
    While edge representations encompass both global shape edges and local texture edges, directly guiding the model to learn these representations could result in model's sensitivity to local textures. To address this, we propose \textit{Shape-disrupted Token Shuffling (STS)} module, which generates neural local edge noise by shuffling token positions from another sample, thereby disrupting category-related shape semantics. By randomly interpolating between the target edge features extracted by ETS and the shape-disrupted texture noise generated by STS, we could effectively preserve category-relevant shape information while perturbing category-irrelevant local textures.
}

\vspace{-0.3cm}
\textbf{Shape-disrupted token shuffling.} Motivated by the insight that holistic shapes could be discerned through the relative positions of edge tokens \cite{baker2018deep,tripathi2023edges,islam2021shape}, we introduce texture noise using shape-disrupted samples by shuffling tokens within the samples. For the input representation $z_j \in \mathbb{R}^{N \times C}$, we employ ${\rm Shuffle}(\cdot)$ function in the token dimension:
\begin{equation}
    z_j^{s} = {\rm Shuffle}(z_j).
    \label{Eq:shuffled tokens}
\end{equation}
This operation could generate neural texture noise while eliminating shape-related information \cite{ren2023masked,tripathi2023edges}, thereby applying diverse shape-irrelevant perturbations to the model learning.

\vspace{0.1cm}
\textbf{Token-level augmentation for shape bias.} 
To effectively enhance the shape bias of the model, we employ token-level interpolation with random sample pairs to generate shape-biased samples.
Concretely, given an input batch $\{z_i\}_{i=1}^B$, where $B$ denotes the batch size, for the $i$-th sample $z_i$, we first extract its edge-pass representation with only edge tokens, \textit{i.e.}, $z_i^{e}$ using Eq.~(\ref{Eq:edge tokens}). Then we randomly select another sample $z_j$ from the batch and generate its token-shuffled version $z_j^{s}$ as Eq.~(\ref{Eq:shuffled tokens}). 
To create shape-biased augmentations, we superimpose the two augmented representations, and each such augmented sample is assigned the label $y_i$ of the sample $z_i^e$ that provides the edge tokens.

Following the Mixup \cite{zhang2018mixup} and CutMix \cite{yun2019cutmix}, we design two interpolation versions of our SETA. 1) For \textit{the Mixup version}, we generate the augmented data $\widetilde{z}_i$ as follows:
\begin{equation} 
    \widetilde{z}_i = \lambda * z_i^{e} + (1 - \lambda) * z_j^{s},
    \label{Eq:SETA-Mixup}
\end{equation}
where $\lambda$ is a random weight sampled from the Beta distribution ${\rm Beta}(\alpha, \beta)$ with $\alpha, \beta \in (0, \infty)$ being two hyper-parameters. 
The Mixup version of SETA can introduce bote style noise and local edge noise into the edge-related tokens, thereby encouraging the model to learn global semantic features from the edge representations.
\textcolor{revised_checked}{2) For \textit{the CutMix version}, we 
generate the token-level binary mask $\mathcal{M}_{\text{edge}}^{\lambda} \in \mathbb{R}^{N \times C}$ with the element being set to $1$ if the corresponding element in $A(\hat{z}_i)$ is in the top $\lambda \sim {\rm Beta}(\alpha, \beta)$ percentage elements. The augmented representation $\widetilde{z}_i$ is:
\begin{equation}
    \widetilde{z}_i = \mathcal{M}_{\text{edge}}^{\lambda} \odot z_i + (1 - \mathcal{M}_{\text{edge}}^{\lambda}) \odot z_j^{s}.
    \label{Eq:SETA-CutMix}
\end{equation}}
The CutMix version of SETA randomly selects a number of edge-related tokens from the original representation and fill the remaining positions with noise tokens from shape-disrupted sample representations, which enforces the model to learn the global dependencies among edge-related tokens.

\begin{figure}[tb!]
    \begin{center}
    \includegraphics[width=\linewidth]{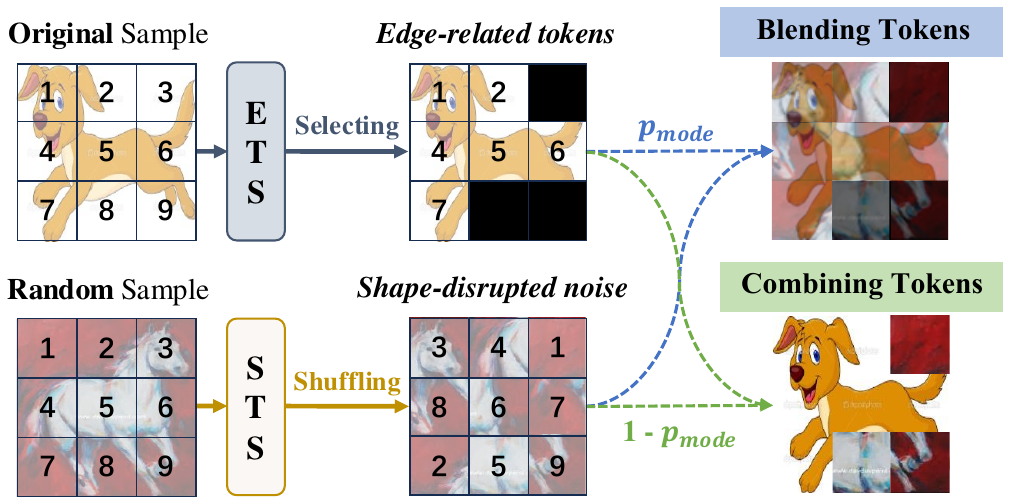}
    \end{center}
    \vspace{-0.2cm}
    \caption{\textcolor{revised_r1}{An intuitive illustration of our SETA. The ETS module extracts edge-related tokens from the original sample, while the STS module generates shape-disrupted noise by randomly selecting another sample from the current batch. The \textit{Mixup} version of SETA blends the values of the edge-related tokens and shape-disrupted tokens, while the \textit{CutMix} version of SETA replaces the edge-irrelated tokens by the shape-disrupted tokens.}}
    \label{fig:idea illustration}
    \vspace{-0.3cm}
  \end{figure}

In practice, for each mini-batch, we randomly select half of the samples to conduct the SETA augmentation. 
\textcolor{revised_r1}{
Within this process, the Mixup variant of SETA is applied with a probability of $p_{\text{mode}}$, while in other cases, the CutMix version of SETA is employed. 
An intuitive illustration of SETA is presented in Fig.~\ref{fig:idea illustration}.
By introducing neural noise from random samples into original samples, domain-specific local edges are perturbed, while category-related shape information is preserved. This allows SETA to encourage the model to learn domain-invariant shape representations, thereby enhancing model generalization.
}
During inference, SETA is closed as conventional augmentation-based DG methods  \cite{li2022uncertainty,zhou2023mixstyle}.

\textbf{Two stylized variants of SETA.}
To adequately enhance model robustness to domain shifts, we further explore two stylized versions of SETA. 
As our SETA primarily relies on texture noise extracted from shape-disrupted neural samples, the diversity of styles within the augmented samples is still limited. 
To address this limitation, we have incorporated two SOTA DG style augmentation methods and devised two variants of SETA, specifically \textit{SETA-S-DSU} (combined with DSU \cite{li2022uncertainty}), which introduces statistical diversity, and \textit{SETA-S-ALOFT} (combined with ALOFT \cite{guo2023aloft}), which introduces frequency diversity.
In this context, when generating token-shuffled samples with disrupted shapes, we also apply channel-level style augmentations (\textit{i.e.}, DSU or ALOFT) to these samples.
The stylized variants could concurrently amplify the diversity of both style and texture edges, thus providing a more comprehensive simulation of potential domain shifts.

In summary, we propose a token-level feature augmentation designed to encourage the shape bias of the model by introducing local edge noise. 
\textcolor{revised_checked}{
Note that our SETA diverges significantly from prior augmentation-based DG works \cite{guo2023aloft,li2022uncertainty,zhou2023mixstyle,zhang2022exact}, which solely perturb style information while keeping other features unchanged.
Since these existing methods ignore spurious local edges within the remaining features, the model inevitably learns domain-specific features and fails to capture shape information effectively.
In contrast, our method proposes to interpolate edge-pass shape-preserved representations from the original image with shape-disrupted local edge representations from another image. 
The method incentivizes not only edge detection but also shape sensitivity: 1) classifying the edge tokens requires the model to exploit category-related edge information; 2) distinguishing the category of edge tokens from the mixed shuffled sample requires the model to distinguish the overall object shape from the shuffled sample edges. 
As a result, \textit{our method could enhance the model learning of class-relevant shape edges while suppressing harmful category-irrelevant local edges and style information.}
Besides, the computational overhead of SETA is minimal, only involving a few matrix operations.
The components within SETA introduce no additional parameters or inference time, resulting in negligible training overhead, \textit{e.g.}, increasing only $0.6\%$ ($0.154s$ vs. $0.153s$) in time for each training step from the baseline on the GFNet-H-Ti architecture.}

\subsection{Theoretical Insight of SETA} 
\label{Theoretical analysis}
We here provide a theoretical insight into SETA and demonstrate that incentivizing model sensitivity to shape features can reduce the generalization risk bound of the model.
\vspace{0.1cm}

\noindent\textbf{Notations.} 
Given a training set of $K$ observed source domains $\mathcal{D}_s = \{D_s^1, D_s^2, ..., D_s^K\}$, the $k$-th domain $D_s^k$ ($1 \leq k \leq K$) contains $n_k$ labeled samples $\{(x_i^k, y_i^k)\}_{i=1}^{n_k}$, where 
$(x_i^k, y_i^k)$ denotes the sample-label pair for the $i$-th sample within the $k$-th domain. 
Let $\mathcal{X}$ and $\mathcal{Y}$ denote the data and label spaces, respectively.
Consider a hypothesis $h: \mathcal{X} \rightarrow \mathcal{Y}$ coming from the hypothesis space $\mathcal{H}$. 
The distributions of the source and target domains are denoted as $\mathcal{D}_s$ and $\mathcal{D}_t$, respectively. 
The loss function is represented by $\ell(\cdot)$, and $d_{\mathcal{H}\Delta \mathcal{H}}[\cdot,\cdot]$ is utilized as the $\mathcal{H}\Delta \mathcal{H}$-divergence \cite{ben2010theory} for quantifying the distribution discrepancy between two domains. 
Denoting the augmentation space as $\mathcal{G}$, an augmentation function $g(\cdot)$ is defined as $g \in \mathcal{G}: \mathcal{X} \rightarrow \mathcal{X}$, which generates augmented data with specific operations.
Since the data augmentation $g(\cdot)$ actually changes the distribution of source domains $\mathcal{D}_s$, the augmented distribution is denoted as $\mathcal{D}_s^{aug}$, and the model trained on this augmented data distribution is represented as $h_{aug}$.
Let $\Pr(\cdot)$ denote the probability of an event, and $\textbf{I}[a]$ serves as the indicator function that yields $1$ if $a$ is true and $0$ otherwise.
\vspace{0.1cm}

To estimate the distribution divergence between two domains, follow \cite{ben2010theory}, we introduce the $\mathcal{H}\Delta \mathcal{H}$-divergence as:
\begin{equation}
    \begin{aligned}
        d_{\mathcal{H}\Delta \mathcal{H}}(\mathcal{D}_s, \mathcal{D}_t) = 2\sup_{h \in \mathcal{H}\Delta \mathcal{H}} \lvert \Pr_{\mathcal{D}_s} (\textbf{I}(h)) - \Pr_{\mathcal{D}_t} (\textbf{I}(h)) \rvert.
    \end{aligned}
    \label{eq:distance definition}
\end{equation}

Given that $D_s$ and $D_t$ are samples that are drawn from $\mathcal{D}_s$ and $\mathcal{D}_t$, the empirical $\mathcal{H}\Delta \mathcal{H}$-divergence estimated on $D_s$ and $D_t$ could be denoted as $d_{\mathcal{H}\Delta \mathcal{H}}(D_s, D_t)$. 
With the sample-label pair $(x, y)$, we use $z$ to denote the feature representation of $(x, y)$ learned by the network. 
Let $\mathcal{Z}$ be the feature space, we redefine the hypothesis $h$ as $h: \mathcal{Z} \rightarrow \mathcal{Y}$, which takes $z$ as input and outputs prediction label. 
For simplicity, we overload $D$ to be the set of feature-label pairs $\{(z, y)\}$.
The risk function of $h$ on domain $D$ is defined as:
\begin{equation}
    R(h; D) = \mathbb{E}_{\langle z, y \rangle \sim D}[\ell(h(z), y)],
\end{equation}
where $\ell(\cdot)$ is the pre-defined loss term. 
Based on the above definitions, we first introduce the generalization error bound \cite{ben2010theory} of the model trained on clean data of source domain $D_s$.

\vspace{0.1cm}
\noindent\textbf{Lemma 1 \cite{ben2010theory}.} 
\textit{Let $\mathcal{H}$ be a hypothesis space of $VC$ dimension $VC(\mathcal{H})$. 
If $D_s$ and $D_t$ are samples of size $M$ each, which are drawn independently from $\mathcal{D}_s$ and $\mathcal{D}_t$, respectively, then for any $\delta \in (0, 1)$ with probability of at least $1 - \delta$ over the random choice of the samples, for all $h \in \mathcal{H}$, 
\begin{equation}
    \begin{aligned}
        R(h; D_t) & \leq R(h; D_s) + \frac{1}{2} d_{\mathcal{H}\Delta \mathcal{H}}(D_s, D_t) \\ 
        & + 4\sqrt{\frac{2 VC(\mathcal{H}) \log(2 M) + \log(2/\delta)}{M}} + \lambda,
    \end{aligned}
\end{equation}
where $\lambda$ is the combined error of the ideal hypothesis $h^*$ that minimizes $R(h; D_s) + R(h; D_t)$.
}
\vspace{0.1cm}

To investigate the effect of feature augmentation on the generalization error bound, we extend the Lemma $1$ to make it applicable to models with feature augmentation.
Specifically, for the souce domain $D_s$ and its augmented data $D_s^{aug}$, noting $D_s \subseteq D_s^{aug}$ and $D_s^{aug}$ could be more difficult to learn than $D_s$, we assume that for the model $h_{aug}$ trained on augmented data, $R(h_{aug}; D_s^{aug})$ is larger than $R(h_{aug}; D_s)$. 
Then we can derive the following theorem:

\vspace{0.1cm}
\noindent\textbf{Theorem 1 (Generalization Risk Bound).} 
\textit{With previous settings, let $\mathcal{H}$ be a hypothesis space of $VC$ dimension $VC(\mathcal{H})$. 
If $D_s^{aug}$ and $D_t$ are samples of size $M$ drawn independently from $\mathcal{D}_s^{aug}$ and $\mathcal{D}_t$, respectively. Suppose that
\begin{equation}
    R(h_{aug} ; D_s^{aug}) - R(h_{aug} ; D_s) \leq \varepsilon(D_s, D_s^{aug}),
\end{equation}
where $R(\cdot; \cdot)$ is the risk function defined as:
\begin{equation}
    R(h; D) = \mathbb{E}_{\langle z, y \rangle \sim D}[\ell(h(z), y)]. 
    \label{eq:risk 1}
\end{equation}
Then for any $\delta \in (0, 1)$ with probability of at least $1 - \delta$, for all $h \in \mathcal{H}$, the following inequality holds:
\begin{equation}
    \begin{aligned}
        |R&(h_{aug}; D_t) - R(h_{aug}; D_s)| \\
        &\leq \varepsilon (D_s, D_s^{aug}) + \frac{1}{2} d_{\mathcal{H}\Delta \mathcal{H}}(D_s^{aug}, D_t) + \zeta + \lambda,
    \end{aligned}
    \label{eq:bound}
\end{equation}
where 
$\zeta=4\sqrt{\frac{2 VC(\mathcal{H}) \log(2 M) + \log(2/\delta)}{M}}$, and 
$\lambda$ is the combined error of the ideal hypothesis $h^*$ on $D_s^{aug}$ and $D_t$. 
}

\noindent\textbf{Proof.} 
Given the source dataset $D_s$, feature augmentation creates diverse data to augment the training data, denoted as $D_s^{aug}$. 
Recall that Lemma $1$ provides a generalization error bound between two different distributions. For $h_{aug}$ trained on the augmented source dataset $D_s^{aug}$, the following inequality holds with a probability of at least $1 - \delta$:
\begin{equation}
    \begin{aligned}
        R(h_{aug}; D_t) \leq &R(h_{aug}; D_s^{aug}) \\
        + &\frac{1}{2} d_{H\Delta H}(D_s^{aug}, D_t) + \zeta + \lambda.
    \end{aligned}
    \label{eq:aug generalization 1}
\end{equation}
Considering that feature augmentation introduces additional perturbations into the model \cite{dao2019kernel,liu2023understanding}, which could potentially enhance the learning difficulty of the model, it is reasonable to assume that \textit{feature augmentation could enlarge the prediction loss of the model} \cite{he2019data,huang2020self}, formulated as:
\begin{equation}
    \begin{aligned}
        R(h_{aug} ; D_s^{aug}) \leq R(h_{aug} ; D_s) + \varepsilon(D_s, D_s^{aug}).
    \end{aligned}
\end{equation}
Therefore, the inequality in Eq.~(\ref{eq:aug generalization 1}) can be reformulated by:
\begin{equation}
    \begin{aligned}
    R(h_{aug}; D_t) \leq &R(h_{aug}; D_s) + \varepsilon (D_s, D_s^{aug}) \\ 
    + &\frac{1}{2} d_{H\Delta H}(D_s^{aug}, D_t) + \zeta + \lambda.
    \end{aligned}
\end{equation}
Rearranging these terms leads to the Theorem $1$. $\hfill \blacksquare$
\vspace{0.1cm}

Theorem $1$ indicates that the upper bound of generalization risk on target domain depends primarily on three terms: 1) $\varepsilon (D_s, D_s^{aug})$ that denotes the consistency between the risk of the model on clean and augmented data; 2) $d_{\mathcal{H}\Delta \mathcal{H}}(D_s^{aug}, D_t)$ that presents the distance between augmented source domains and target domain; 
3) $\lambda$ that is the combined error of the optimal hypothesis $h^*$ on all domains. 
These terms are all influenced by two factors: 1) the model's ability to extract invariant information between source and target domains; and 2) the augmentation to reduce the discrepancy between these domains.
Regarding the first factor, 
we analyze it from the architectural perspective.
For CNN models, which heavily rely on convolutions with limited receptive field, the feature response of convolution can be expressed as $\Phi_{\text{CNN}}(x) = W_{\text{conv}} * z$, where $W_{\text{conv}}$ is convolutional kernel, and $*$ is convolution operation. 
Due to the high sensitivity of convolution to local textures, significant differences in $z$ between domains (\textit{e.g.}, varying textures) can lead to an increased discrepancy in $\Phi_{\text{CNN}}(z)$ between source and target domains. This increase contributes to a larger $d_{\mathcal{H}\Delta \mathcal{H}}(D_s^{aug}, D_t)$ and a greater $\lambda$. 
Therefore, CNNs perform inferiorly in domains with substantial gaps. 
In contrast to CNNs, token-based models (ViTs and MLPs) learn global dependencies among tokens, focusing more on global shape information rather than local textures. 
As a result, they are better at capturing shape information from different domains, thereby reducing domain gap \( d_{H\Delta H}(D_s, D_t) \) and achieving a smaller $\lambda$ of the optimal hypothesis $h^*$. 
The above analysis indicates that the high dependency of CNNs on local convolution makes them more susceptible to domain shifts, whereas ViTs and MLPs, with their learning of global token dependencies, exhibit superior generalization ability. 
Therefore, using advanced architectures can effectively reduce the model's generalization error bound.

Besides, we analyze the effects of feature augmentation on generalization error bound. 
Since the augmentation $g(\cdot)$ could potentially affect both the two terms by changing the distribution of source domains, we here investigate how to design $g(\cdot)$ for reducing the two terms. 
To simplify the question, following \cite{dao2019kernel,he2019data}, we consider a feature extractor $\textbf{h}$ with a linear softmax classifier $\textbf{W} = [\textbf{w}_1, \textbf{w}_2, ..., \textbf{w}_L]\in \mathbb{R}^{C \times L}$, where $C$ is the number of channels, and $L$ is the number of categories. 
For augmentation $g \sim \mathcal{G}$, training risk on augmented data is:
\begin{equation}
    R_{aug} = \frac{1}{N} \sum_{n=1}^N \mathbb{E}_{g \sim \mathcal{G}}[\ell(\textbf{W}^\top \textbf{h}(g(z)), y)].
\end{equation}
\textcolor{revised_checked}{Following previous works \cite{tripathi2023edges,hermann2020shapes,shi2020informative} that emphasize shape features for model generalization, we assume that the extracted features $\textbf{h}$ could be decomposed to \textit{global shape features} $h_{gs}$, \textit{local edge features} $h_{le}$ and \textit{style features} $h_{st}$. }
Considering that the model is trained to optimize the loss $R_{aug}$ on augmented source data, we analyze the impacts of $g(\cdot)$ on the optimization process and derive Proposition $1$.

\noindent\textbf{Proposition 1 (Augmentation for Generalization).} 
\textit{
    To reduce the $\varepsilon(D_s, D_s^{aug})$ and $d_{\mathcal{H}\Delta \mathcal{H}}(D_s^{aug}, D_t)$ when a large distribution gap of clean and augmented data exists, the augmentation function $g(\cdot)$ should adhere to:
    \begin{equation}
        \left\{
            \begin{aligned}
                &|h_{gs}(g(z)) - h_{gs}(z)| < \xi, \\
                &|h_{le}(g(z)) - h_{le}(z)| > \epsilon_1, \\
                &|h_{st}(g(z)) - h_{st}(z)| > \epsilon_2, \\
            \end{aligned}
        \right. \\
    \end{equation}
    where $\xi > 0$ is a small value, and $\epsilon_1, \epsilon_2 \geq \xi$.
    The solution requires the weights $w_{le} \rightarrow 0$ and $w_s \rightarrow 0 $, regularizing the model to learn shape features $h_{gs}$ for better generalization.
}

\noindent\textbf{Proof.} To investigate how to design $g(\cdot)$ for reducing both $\varepsilon(D_s, D_s^{aug})$ and $d_{\mathcal{H}\Delta \mathcal{H}}(D_s^{aug}, D_t)$, following \cite{dao2019kernel,he2019data}, we explore the simplified problem in conjunction with a linear softmax classifier $\textbf{W} = [\textbf{w}_1, \textbf{w}_2, ..., \textbf{w}_L]\in \mathbb{R}^{C \times L}$, where $C$ is the number of channels, and $L$ is the number of categories. 
Let $\textbf{h}(z) = [h_1(z), h_2(z), \dots, h_C(z)]^\top$ denote feature vector extracted by the model.
Given the augmentation function $g \in \mathcal{G}: \mathcal{Z} \rightarrow \mathcal{Z}$, we reformulate the risk $R(h_{aug}; D_s^{aug})$ in Eq.~(\ref{eq:risk 1}) as:
\begin{equation}
    R(h_{aug}; D_s^{aug}) = \frac{1}{N_s} \sum_{n=1}^{N_s} \mathbb{E}_{g \sim \mathcal{G}}[\ell(\textbf{W}^\top \textbf{h}(g(z_s)), y_s)].
\end{equation}
which can be expanded using the Taylor expansion \cite{dao2019kernel} to:
\begin{equation}
    \begin{aligned}
    \mathbb{E}_{g \sim \mathcal{G}}[\ell(\textbf{W}^\top \textbf{h}(g(z_s)), y_s)&] = \ell(\textbf{W}^\top \overline{\textbf{h}}, y_s) + \\
    &\frac{1}{2} \mathbb{E}_{g \sim \mathcal{G}}[\bigtriangleup^\top \textbf{H}(\tau, y_s)\bigtriangleup],
    \end{aligned}
    \label{eq:Taylor}
\end{equation}
where $\overline{\textbf{h}}=\mathbb{E}_{g \sim \mathcal{G}}[\textbf{h}(g(z))]$ and $\bigtriangleup=\textbf{W}^\top (\overline{\textbf{h}}-\textbf{h}(g(z)))$. 
Notably, the Hessian matrix $\textbf{H}$ is positive semi-definite and independent of $y$ for cross-entropy loss with softmax \cite{dao2019kernel}. 
When optimizing the empirical loss $R(h_{aug}; D_s^{aug})$, the second-order term in Eq.~(\ref{eq:Taylor}) works as a regularization term, \textit{i.e.}, requiring the weight $w \rightarrow 0$ if the variance of its corresponding feature $h(g(z))$ is large.

To narrow the generalization error bound in Eq.~(\ref{eq:bound}), we aim to reduce the domain gap $d_{\mathcal{H}\Delta \mathcal{H}}(D_s^{aug}, D_t)$ while keeping $\varepsilon(D_s, D_s^{aug})$ small.
According to Eq.~(\ref{eq:distance definition}), since the predicted label of the model can be denoted as $\textbf{W}^\top \textbf{h}$, we reformulate the domain discrepancy $d_{\mathcal{H}\Delta \mathcal{H}}(D_s^{aug}, D_t)$ as:
\begin{equation}
    \begin{aligned}
        d_{H\Delta H}(D_{s}^{aug}, D_t) = 2 \sup_{h \in \mathcal{H} \Delta \mathcal{H}} & \lvert \mathbb{E}_{z \sim D_s}[\mathbf{I}(\textbf{W}^\top \textbf{h}(g(z)))] \\ & - \mathbb{E}_{z \sim D_t}[\mathbf{I}(\textbf{W}^\top \textbf{h}(z))] \rvert 
    \end{aligned}
\end{equation}
Denoting the features being invariant across source and target domains as $h_{di}$, and the features being specific for source domains as $h_{ds}$.
The above equation indicates that to reduce the domain gap, we should constrain the model to enhance the learning of the weight $w_{di}$ corresponding to $h_{di}$ while reducing the learning of $w_{ds}$ corresponding to $h_{ds}$. 
\textcolor{revised_checked}{
Inspired by pioneering works \cite{tripathi2023edges,hermann2020shapes,shi2020informative} that divides features into style, texture, and shape, we formulate the hypothesis specifically for DG tasks, assuming that the extracted features $\textbf{h}$ could be decomposed to \textit{global shape features} $h_{gs}$, \textit{local edge features} $h_{le}$ and \textit{style features} $h_{st}$. 
The local edge features $h_{le}$ and style features $h_{st}$ mainly contain domain-specific information, while the global shape features $h_{gs}$ mainly encode domain-invariant information.} 
To reduce $\varepsilon(D_s, D_s^{aug})$ and $d_{\mathcal{H}\Delta \mathcal{H}}(D_s^{aug}, D_t)$, the augmentation $g(\cdot)$ should adhere to:
\begin{equation}
    \left\{
        \begin{aligned}
            &|h_{gs}(g(z)) - h_{gs}(z)| < \xi, \\
            &|h_{le}(g(z)) - h_{le}(z)| > \epsilon_1, \\
            &|h_{st}(g(z)) - h_{st}(z)| > \epsilon_2, \\
        \end{aligned}
    \right. \\
\end{equation}
where $\xi > 0$ is a small value, and $\epsilon_1, \epsilon_2 \geq \xi$. 

The inequalities necessitate that the function $g(\cdot)$ induces significant perturbations to both $h_{le}$ and $h_{st}$, thereby resulting in substantial variations in domain-specific features.
Recalling that the second term in Eq.~(\ref{eq:Taylor}) serves as a regularization term. 
This term restricts the weights $w_{le} \rightarrow 0$ and $w_{st} \rightarrow 0$, thus suppressing the model learning of domain-specific features. 
Simultaneously, due to minimal changes in the global shape features $h_{gs}$, the regularization term promotes the learning of the corresponding weight $w_{ge}$, 
thus enhancing the model learning of domain-invariant features. 
Consequently, the domain gap $d_{\mathcal{H}\Delta \mathcal{H}}(D_s^{aug}, D_t)$ could be effectively reduced during training.
Furthermore, given that shape features primarily encode semantic information crucial for classification, the consistency $\varepsilon(D_s, D_s^{aug})$ could also be decreased.
$\hfill \blacksquare$
\label{sec:theorem_and_proposition}

The above inequalities highlight the effectiveness of our SETA, which primarily perturbs local edge features $h_{le}$ and style features $h_{st}$ while preserving shape features $h_{gs}$.
\textcolor{revised_checked}{
As a result, \textit{SETA could increase $\epsilon_1$ and $\epsilon_2$ while keeping $\xi$ small, thus effectively diminishing both the $\varepsilon(D_s, D_s^{aug})$ and $d_{\mathcal{H}\Delta \mathcal{H}}(D_s^{aug}, D_t)$ during training. }
    \textbf{Theorem 1} and \textbf{Proposition 1} jointly analyze the model's generalization error bound from both the architectural and augmentation perspectives. 
    Existing DG methods, which are predominantly based on CNNs, suffer the texture bias due to the limited receptive field of local convolutions, resulting in larger $d_{\mathcal{H}\Delta \mathcal{H}}(D_s^{aug}, D_t)$ and $\lambda$.
    Besides, previous DG augmentation methods \cite{guo2023aloft,li2022uncertainty,zhou2023mixstyle} are primarily designed for CNNs, focusing on perturbing channel-level style information while ignoring to suppress spatial-level local spurious edges. 
    Due to the lack of explicit constraints to learn shape information, these methods may inadvertently encourage the model to excessively learn domain-specific local edge information, leading to larger $\varepsilon(D_s, D_s^{aug})$ and $d_{\mathcal{H}\Delta \mathcal{H}}(D_s^{aug}, D_t)$.
These issues increase generalization error bound of existing CNN-based methods.
In contrast, \textit{our proposed method delves into local edge perturbations to increase $\epsilon_1$, potentially compelling the model to focus on shape representations}. The two stylized variants of SETA could further increase the $\epsilon_1$ and $\epsilon_2$, thereby effectively reducing the generalization error bound. }
The empirical evidence in Sec.~\ref{sec:domain gap} supports the assertion that our method can significantly enhance the model learning of shape information while reducing the domain gap between source and target domains.

\section{Experiment}

\subsection{Datasets}
We conduct comprehensive experiments on the DomainBed benchmark \cite{gulrajani2020search} and evaluate our methods on five conventional DG datasets: 
(1) \textbf{PACS} \cite{li2017deeper} consists of $9,991$ images of $7$ categories from $4$ domains: Photo, Art Painting, Cartoon, and Sketch. 
(2) \textbf{Office-Home} \cite{venkateswara2017deep} contains $15,588$ images of $65$ categories from $4$ domains: Artistic, Clipart, Product and Real-World. 
(3) \textbf{VLCS} \cite{torralba2011unbiased} comprises $10,729$ images of $5$ categories selected from $4$ domains, VOC $2007$ (Pascal), LabelMe, Caltech and Sun. 
(4) \textbf{Terra Incognita} \cite{beery2018recognition} contains photographs of wild animals taken by $4$ camera-trap domains, with $10$ classes and a total of $24,788$ images. 
(5) \textbf{DomainNet} \cite{peng2019moment} is a large-scale dataset, consisting of $586,575$ images with $345$ categories from $6$ domains, \textit{i.e.}, Clipart, Infograph, Painting, Quickdraw, Real, and Sketch.
\subsection{Implementation Details}

\renewcommand{\thefootnote}{\fnsymbol{footnote}} \begin{table*}[tb!]
    \centering
    \caption{
    Performance (\%) comparisons with SOTA DG methods on DomainBed benchmark. We conduct a comprehensive evaluation by comparing our methods with $15$ existing DG algorithms across five diverse multi-domain datasets: PACS \cite{li2017deeper}, VLCS \cite{torralba2011unbiased}, OfficeHome \cite{venkateswara2017deep}, TerraInc \cite{beery2018recognition}, and DomainNet \cite{peng2019moment}. 
    The CNN-based methods in our experiments employ the ResNet-50 architecture, and we select two SOTA transformer-based models with similar parameter amounts to ResNet-$50$, \textit{i.e.}, Swin-T \cite{liu2021swin} and GFNet-H-Ti \cite{rao2021global} for our methods. 
    We use the validation set from source domains for model selection. The best is \textbf{bold} while the second-best is \underline{underlined}. 
    }
    \vspace{-0.2cm}
    \resizebox{\linewidth}{!}{
    \begin{tabular}{l|c|ccccc|c}
          \hline
        Method & Backbone & PACS & VLCS & OfficeHome & TerraInc & DomainNet & \textbf{Avg.} \\
        \hline
        Baseline \cite{gulrajani2020search} {\scriptsize (ICLR'20)} & ResNet50 & 85.50 $\pm$ 0.20 & 77.50 $\pm$ 0.40 & 66.50 $\pm$ 0.30 & 46.10 $\pm$ 1.80 & 40.90 $\pm$ 0.10 & 63.30 \\
        Mixup \cite{zhang2018mixup} {\scriptsize (ICLR'18)} & ResNet50 & 84.60 $\pm$ 0.60 & 77.40 $\pm$ 0.60 & 68.10 $\pm$ 0.30 & 47.90 $\pm$ 0.80 & 39.20 $\pm$ 0.10 & 63.40 \\
        RSC \cite{huang2020self} (ECCV'20) & ResNet50 & 85.20 $\pm$ 0.90 & 77.10 $\pm$ 0.50 & 65.50 $\pm$ 0.90 & 46.60 $\pm$ 1.00 & 38.90 $\pm$ 0.50 & 62.70 \\
        SagNet \cite{nam2021reducing} {\scriptsize (CVPR'21)} & ResNet50 & 86.30 $\pm$ 0.20 & 77.80 $\pm$ 0.50 & 68.10 $\pm$ 0.10 & 48.60 $\pm$ 1.00 & 40.30 $\pm$ 0.10 & 64.20 \\
        FISH \cite{shi2021gradient} {\scriptsize (ICLR'21)} & ResNet50 & 85.50 $\pm$ 0.30 & 77.80 $\pm$ 0.30 & 68.60 $\pm$ 0.40 & 45.10 $\pm$ 1.30 & 42.70 $\pm$ 0.20  & 63.90 \\
        SWAD \cite{cha2021swad} {\scriptsize (NeurIPS'21)} & ResNet50 & 88.10 $\pm$ 0.10 & 79.10 $\pm$ 0.10 & 70.60 $\pm$ 0.20 & 50.00 $\pm$ 0.30 & 46.50 $\pm$ 0.10 & 66.90 \\
        GVRT \cite{min2022grounding} {\scriptsize (ECCV'22)} & ResNet50 & 85.10 $\pm$ 0.30 & 79.00 $\pm$ 0.20 &  70.10 $\pm$ 0.10 & 48.00 $\pm$ 0.20 & 44.10 $\pm$ 0.10 & 65.20 \\
        MIRO \cite{cha2022domain} {\scriptsize (ECCV'22)} & ResNet50 & 85.40 $\pm$ 0.40 & 79.00 $\pm$ 0.00 & 70.50 $\pm$ 0.40 & 50.40 $\pm$ 1.10 & 44.30 $\pm$ 0.20 & 65.90 \\
        PTE \cite{min2022grounding} {\scriptsize (ECCV'22)} & ResNet50 & 85.10 $\pm$ 0.30 & 79.00 $\pm$ 0.20 & 70.10 $\pm$ 0.10 & 48.00 $\pm$ 0.20 & 44.10 $\pm$ 0.10 & 65.20 \\
        MVRML \cite{zhang2022mvdg} {\scriptsize (ECCV'22)} & ResNet50 & 88.30 $\pm$ 0.30 & 77.90 $\pm$ 0.20 & 71.30 $\pm$ 0.10 & 51.00 $\pm$ 0.60 & 45.70 $\pm$ 0.10 & 66.80 \\
        EQRM \cite{eastwood2022probable} {\scriptsize (NeurIPS'22)} & ResNet50 & 86.50 $\pm$ 0.20 & 77.80 $\pm$ 0.60 & 67.50 $\pm$ 0.10 & 47.80 $\pm$ 0.60 & 41.00 $\pm$ 0.30 & 64.10 \\
        DAC-SC \cite{lee2023decompose} {\scriptsize (CVPR'23)} & ResNet50 & 87.50 $\pm$ 0.10 & 78.70 $\pm$ 0.30 & 70.30 $\pm$ 0.20 & 44.90 $\pm$ 0.10 & 46.50 $\pm$ 0.30 & 65.60 \\
        CCFP \cite{li2023cross} {\scriptsize (ICCV'23)} & ResNet50 & 86.60 $\pm$ 0.20 & 78.90 $\pm$ 0.30 & 68.90 $\pm$ 0.10 & 48.60 $\pm$ 0.40 & 41.20 $\pm$ 0.00 & 64.80 \\
        DomainDrop \cite{guo2023domaindrop} {\scriptsize (ICCV'23)} & ResNet50 & 87.90 $\pm$ 0.30 & 79.80 $\pm$ 0.30 & 68.70 $\pm$ 0.10 & 51.50 $\pm$ 0.40 & 44.40 $\pm$ 0.50 & 66.50 \\
        \hline
        \hline
        Mixup \cite{zhang2018mixup} {\scriptsize (ICLR'18)} & Swin-T & 86.95 $\pm$ 0.47 & 78.99 $\pm$ 0.23 & 73.42 $\pm$ 0.02 & 51.95 $\pm$ 0.12 & 45.34 $\pm$ 0.11 & 67.33 \\
        CutMix \cite{yun2019cutmix} {\scriptsize (ICCV'19)} & Swin-T & 84.91 $\pm$ 0.34 & 78.93 $\pm$ 0.25 & 73.47 $\pm$ 0.22 & 51.98 $\pm$ 0.29 & 45.41 $\pm$ 0.07 & 66.94 \\
        MixStyle \cite{zhou2023mixstyle} {\scriptsize (IJCV'22)} & Swin-T & 86.67 $\pm$ 0.20 & 78.85 $\pm$  0.21 & 72.83 $\pm$ 0.30 & 50.67 $\pm$ 0.57 & 40.63 $\pm$ 0.16 & 65.93 \\
        DSU \cite{li2022uncertainty} {\scriptsize (ICLR'22)} & Swin-T & 86.98 $\pm$ 0.44 & 79.06 $\pm$ 0.23 & 73.34 $\pm$ 0.15 & 51.66 $\pm$ 0.25 & 45.42 $\pm$ 0.19 & 67.29 \\
        ALOFT \cite{guo2023aloft} {\scriptsize (CVPR'23)} & Swin-T & 86.74 $\pm$ 0.21 & 79.18 $\pm$ 0.60 & 73.41 $\pm$ 0.13 & 52.11 $\pm$ 0.36 & 45.47 $\pm$ 0.06 & 67.38 \\
        \hline
        Baseline \cite{touvron2021training} {\scriptsize (ICML'21)} & Swin-T & 86.27 $\pm$ 0.71 & 78.57 $\pm$ 0.54 & 73.25 $\pm$  0.15 & 51.33 $\pm$ 0.28  & 45.17 $\pm$ 0.20 & 66.92 \\
        \rowcolor{mygray} \textbf{SETA {\scriptsize (Ours)}} & Swin-T & 87.46 $\pm$ 0.42 & 79.34 $\pm$ 0.13 & 73.61 $\pm$ 0.20 & 51.52 $\pm$ 0.42 & 45.57 $\pm$ 0.03 & 67.50 \\
        \rowcolor{mygray} \textbf{SETA-S-DSU {\scriptsize (Ours)}} & Swin-T & 87.56 $\pm$ 0.20 & 79.52 $\pm$ 0.53 & \underline{73.85 $\pm$ 0.09} & 52.51 $\pm$ 0.41 & 45.63 $\pm$ 0.04 & 67.81 \\
        \rowcolor{mygray} \textbf{SETA-S-ALOFT {\scriptsize (Ours)}} & Swin-T & 87.83 $\pm$ 0.13 & 79.80 $\pm$ 0.44 & \textbf{73.87 $\pm$ 0.11} & \textbf{52.87 $\pm$ 0.98} & 45.64 $\pm$ 0.22 & 68.00 \\
        \hline
        \hline
        Mixup \cite{zhang2018mixup} {\scriptsize (ICLR'18)} & GFNet-H-Ti & 88.75 $\pm$ 0.22 & 79.23 $\pm$ 0.23 & 71.50 $\pm$ 0.19 & 51.22 $\pm$ 0.62 &  44.26 $\pm$ 0.05 & 66.99 \\
        CutMix \cite{yun2019cutmix} {\scriptsize (ICCV'19)} & GFNet-H-Ti & 88.79 $\pm$ 0.43 & 79.25 $\pm$ 0.09 & 71.54 $\pm$ 0.36 & 51.64 $\pm$ 0.82 & 44.38 $\pm$ 0.08 & 67.12 \\
        MixStyle \cite{zhou2023mixstyle} {\scriptsize (IJCV'22)} & GFNet-H-Ti & 88.81 $\pm$ 0.40 & 78.95 $\pm$ 0.38 & 71.40 $\pm$ 0.16 & 51.68 $\pm$ 0.93 & 42.65 $\pm$ 0.05 & 66.70 \\
        DSU \cite{li2022uncertainty} {\scriptsize (ICLR'22)} & GFNet-H-Ti & 89.55 $\pm$ 0.30 & 79.04 $\pm$ 0.17 & 71.32 $\pm$ 0.23 & 51.23 $\pm$ 0.25 & 45.52 $\pm$ 0.03 & 67.33 \\
        ALOFT \cite{guo2023aloft} {\scriptsize (CVPR'23)} & GFNet-H-Ti & 89.60 $\pm$ 0.10 & 79.70 $\pm$ 0.80 & 72.10 $\pm$ 0.20 & 52.30 $\pm$ 1.50 & \underline{46.50 $\pm$ 0.10} & 68.04 \\
        \hline
        Baseline \cite{touvron2021training} {\scriptsize (ICML'21)} & GFNet-H-Ti & 87.21 $\pm$ 0.35 & 78.90 $\pm$ 0.30 & 71.11 $\pm$ 0.18 & 50.30 $\pm$ 0.50 & 45.40 $\pm$ 0.10 & 66.59 \\
        \rowcolor{mygray} \textbf{SETA {\scriptsize (Ours)}} & GFNet-H-Ti & 90.52 $\pm$ 0.21 & 79.71 $\pm$ 0.22 & 72.33 $\pm$ 0.13 & 52.56 $\pm$ 0.63 & 45.95 $\pm$ 0.40 & 68.21 \\
        \rowcolor{mygray} \textbf{SETA-S-DSU {\scriptsize (Ours)}} & GFNet-H-Ti & \textbf{90.73 $\pm$ 0.28} & \underline{79.91 $\pm$ 0.15} & 72.51 $\pm$ 0.12 & 52.69 $\pm$ 0.05 & 45.97 $\pm$ 0.32 & \underline{68.36} \\
        \rowcolor{mygray} \textbf{SETA-S-ALOFT {\scriptsize (Ours)}} & GFNet-H-Ti & \underline{90.69 $\pm$ 0.21} & \textbf{79.93 $\pm$ 0.39} & 72.70 $\pm$ 0.10 & \underline{52.85 $\pm$ 0.29} & \textbf{46.73 $\pm$ 0.08} & \textbf{68.58} \\
      \hline
    \end{tabular}
     }
    \label{tab:domainbed}
    \vspace{-0.2cm}
  \end{table*}

\textbf{Basic details.} 
We follow the leave-out-one-domain strategy to evaluate performance for different DG datasets, where one domain is used for testing, and the remaining domains are employed for training.
The data from source domains are split into training subsets ($80\%$) and validation subsets ($20\%$).
For all experiments, we employ the AdamW optimizer and use the default hyperparameters of DomainBed, with the batch size as $32$, the learning rate as $5\times 10^{-4}$ and the weight decay as $0$.
We use the ImageNet-pretrained models as backbones and train the network for $5000$ iterations.
We select the best model on the validation subsets of source domains and report the top-$1$ accuracy for the entire target domain. 
We repeat the entire experiment three times using different seeds and report the mean and standard error over all the repetitions.

\noindent\textbf{Method-specific details.} 
In our experiments, we explore two representative ViT and MLP backbones in our experiments, \textit{i.e.}, the ViT model Swin Transformer (Swin in short) \cite{liu2021swin} and the MLP model GFNet \cite{rao2021global}, which have been widely used in previous DG works \cite{guo2023domaindrop,guo2023aloft,kim2022broad,liao2023domain}.
Specifically, we employ the Swin-T with $27$M parameters, and the GFNet-H-Ti with $14$M parameters, as their parameter counts are either comparable or lower than that of ResNet-50 with $23.5$M parameters.
To determine the optimal values of the method-specific hyperparameters, \textit{i.e.}, $p$ representing the ratio of edge-related token selection, and $p_{\text{mode}}$ indicating the probability of selecting either the Mixup or CutMix version of SETA, we perform a grid search with $p \in \{0.25, 0.33, 0.50, 0.67, 0.75, 0.90, 1.0\}$ and $p_{\text{mode}} \in \{0.2, 0.4, 0.6, 0.8\}$ using the validation set.
Our SETA is applied to all network layers, and following \cite{guo2023domaindrop}, we randomly activate SETA in one layer during each iteration. 
The $\alpha$ and $\beta$ of Beta distribution are set to $4$ and $1$, respectively.
Regarding the stylization modules, \textit{i.e.}, DSU \cite{li2022uncertainty} and ALOFT \cite{guo2023aloft} employed in stylized variants of SETA, we use the same hyperparameters as in their original papers in our experiments.

\subsection{Compared with SOTA Methods}

\textbf{Results on DomainBed.} 
In Tab.~\ref{tab:domainbed}, we present a comprehensive comparison between our SETA method and \textbf{18} recent algorithms for DG on the DomainBed benchmark \cite{gulrajani2020search}. 
We also compare our method with the baseline model (\textit{i.e.}, vanilla ERM \cite{gulrajani2020search}) on the same backbones that have been explored in pioneering DG works \cite{guo2023aloft,kim2022broad,liao2023domain}. 
Specifically, on the Swin-T backbone, our SETA exhibits a notable improvement over the baseline model by a margin of $0.58\%$ on average ($67.50\%$ vs. $66.92\%$), while the two variants of SETA demonstrate further enhancements, \textit{e.g.}, SETA-S-ALOFT surpasses the baseline by $1.08\%$ ($68.00\%$ vs. $66.92\%$).
Furthermore, substantial improvements are observed when switching to the GFNet-H-Ti backbone. SETA enhances the baseline accuracy by $3.31\%$ ($90.52\%$ vs. $87.21\%$) on PACS, $0.81\%$ ($79.71\%$ vs. $78.90\%$) on VLCS, $1.22\%$ ($72.33\%$ vs. $71.11\%$) on OfficeHome, $2.26\%$ ($52.56\%$ vs. $50.30\%$) on TerraInc, and $0.55\%$ ($45.95\%$ vs. $45.40\%$) on DomainNet. 
The variants of our SETA achieve state-of-the-art performances on all datasets, with SETA-S-ALOFT surpassing the baseline on average by $1.99\%$ ($68.58\%$ vs. $66.59\%$), while SETA-S-DSU outperforms the baseline by $1.77\%$ ($68.36\%$ vs. $66.59\%$).

As shown in Tab.~\ref{tab:domainbed}, we observe promising performance from both the Swin-T and GFNet-H-TI backbones. 
Specifically, Swin-T surpasses ResNet-$50$ by $3.62\%$ ($66.92\%$ vs. $63.30\%$), while GFNet-H-Ti outperforms ResNet-50 by $3.29\%$ ($66.59\%$ vs. $63.30\%$). 
These results indicate the superiority of ViT and MLP architectures, attributed to their capability to learn dependencies among tokens \cite{sultana2022self,dosovitskiy2020image}.
Our SETA method can effectively incentivize the shape bias, thereby further enhancing the performance of these advanced networks. 
Notably, SETA significantly outperforms SOTA CNN-based methods, \textit{e.g.}, surpassing the normalization method DAC-SC by $2.61\%$ ($68.21\%$ vs. $65.60\%$) and the dropout-based method DomainDrop by $1.71\%$ ($68.21\%$ vs. $66.50\%$) on GFNet-H-Ti.


\textbf{Comparison with SOTA augmentation methods.} 
    We compare SETA with SOTA augmentation-based DG methods on the same architectures, including the image-level augmentation methods (\textit{i.e.}, Mixup \cite{zhang2018mixup} and CutMix \cite{yun2019cutmix}) and feature-level augmentation methods (\textit{i.e.}, MixStyle \cite{zhou2023mixstyle}, DSU \cite{li2022uncertainty}, and ALOFT \cite{guo2023aloft}). 
    Specifically, we notice that the Mixup and CutMix methods cannot adequately improve the model performance, which is due to the limited diversity of their augmented samples.
    Besides, style augmentation methods, which aim to simulate domain shift by solely perturbing channel-level style features, fail to suppress the spurious edges within the unchanged features. 
    On the contrary, to explicitly promote the shape bias of the model, SETA is designed to perturb spurious local edges during training, thereby notably enhancing the generalization performance.
    Among these methods, SETA consistently demonstrates superior overall performance, outperforming DSU by $0.88\%$ ($68.21\%$ vs. $67.33\%$) and ALOFT by $0.17\%$ ($68.21\%$ vs. $68.04\%$)  on GFNet-H-Ti. 
    Furthermore, we design two stylized variants of SETA by integrating two SOTA style augmentation DG methods, namely SETA-S-DSU employing DSU \cite{li2022uncertainty} and SETA-S-ALOFT employing ALOFT \cite{guo2023aloft}. 
    As showing in Tab.~\ref{tab:domainbed}, both SETA-S-DSU and SETA-S-ALOFT can significantly enhances the performance of existing augmentation methods, \textit{e.g.}, SETA-S-DSU improves DSU by $0.52\%$ ($67.81\%$ vs. $67.29\%$) on Swin-T and $1.03\%$ ($68.36\%$ vs. $67.33\%$) on GFNet-H-Ti. These results prove the orthogonality of our method with existing augmentation methods and emphasize the significance of enhancing shape bias alongside style diversity for boosting model generalization.

\begin{table}[tb!] 
    \begin{center}
  \caption{
      Comparisons (\%) with the SOTA augmentation methods. The experiments are conducted on DeepLab-v$3$ \cite{chen2018encoder} segmentation network with ResNet$50$ as the backbone. We train the model on synthetic GTA$5$ while using real CityScapes, BDD-$100$K, and Mapillary for testing. The best is \textbf{bold}. 
      }
      \vspace{-0.3cm}
      \label{tab: segmentation results}
    \resizebox{\linewidth}{!}{
    \begin{tabular}{l|ccc|c}
        \hline
        Method & CityScapes & BDD-100K & Mapillary & Avg. \\
        \hline
        Baseline \cite{chen2018encoder} {\scriptsize (ECCV'18)} & 28.95 & 25.14 & 28.18 & 27.42 \\
        MixStyle \cite{zhou2023mixstyle} {\scriptsize (ICLR'21)} & 30.68 & 30.09 & 31.13 & 30.63 \\
        EFDM \cite{zhang2022exact} {\scriptsize (CVPR'22)} & 30.10 & 27.93 & 32.28 & 30.10 \\
        DSU \cite{li2022uncertainty} {\scriptsize (ICLR'22)} & 32.01 & 30.58 & 32.05 & 31.55 \\
        \rowcolor{mygray} SETA (Ours) & \textbf{32.84} & \textbf{30.68} & \textbf{33.60} & \textbf{32.37} \\
        \hline
    \end{tabular}}
    \end{center}
    \vspace{-0.1cm}
  \end{table}

\begin{figure}[tb!]
    \centering
    \includegraphics[scale=0.31]{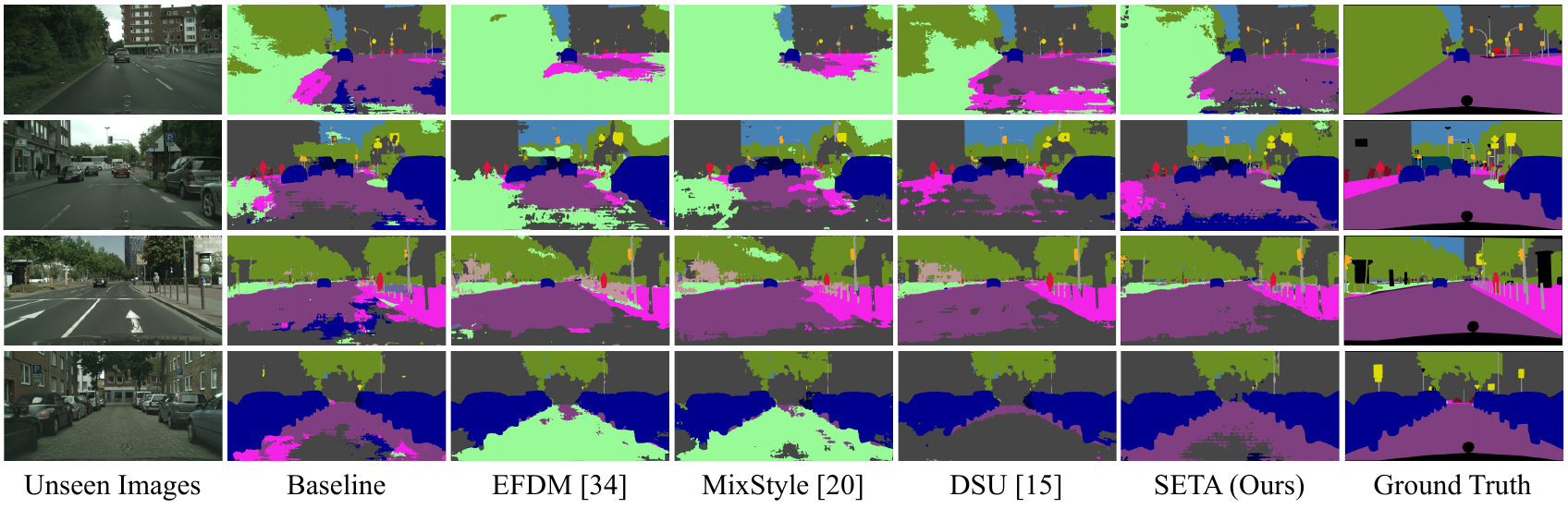}
    \vspace{-0.6cm}
    \caption{
        The segmentation results on unseen domain CityScapes with the model trained on synthetic GTA$5$. We compare our method with the baseline and SOTA augmentation-based DG methods. The results indicate that our method can effectively improve the segmentation performance of the model.
    }
    \vspace{-0.2cm}
    \label{fig: segmentation}
\end{figure}

\subsection{Generalization on Semantic Segmentation}
\textbf{Datasets and implementation details.} 
To evaluate the cross-domain generalization ability of segmentation models, we adopt the synthetic dataset GTA$5$ \cite{richter2016playing} for model training while using the real-world datasets for testing, including CityScapes \cite{cordts2016cityscapes}, BDD-$100$K \cite{yu2020bdd100k}, and Mapillary \cite{neuhold2017mapillary}. 
We follow the previous work \cite{choi2021robustnet} and adopt the DeepLabV$3$+ \cite{chen2018encoder} with ResNet$50$ as the backbone for the semantic segmentation architecture.
The SGD optimizer with an initial learning rate of $1e-2$ and momentum of $0.9$ is utilized. 
For the quantitative experiments, the mean Intersection over Union (mIoU) is used to measure the segmentation performance of the model.

\textbf{Experimental results.} We present the results in Tab.~\ref{tab: segmentation results} and compare our SETA with SOTA augmentation-based DG methods. 
Specifically, we observe that existing augmentation methods can effectively enhance performance by perturbing style statistics, thereby improving the model's robustness to style variations.
However, since these methods ignore potentially spurious local edges in features, the model could still learn domain-specific information and overfit to the source domain. 
Differently, our SETA could effectively suppress spurious edge features, thus enhancing the model learning of semantic structure information.
As shown in Tab.~\ref{tab: segmentation results}, our SETA achieves state-of-the-art overall performance, yielding the baseline by $4.95\%$ ($32.37\%$ vs. $27.42\%$). 
The visualization results in Fig.~\ref{fig: segmentation} further prove the effectiveness of our method.



\subsection{Analytical Experiments}

\begin{table*}[tb!] 
    \begin{center}
    \caption{
        Ablation study (\%) on different modules of SETA on the DomainBed benchmark.
        We denote the Activation-based Edge Tokens Selection as ETS and the Shape-disrupted Token Shuffling as STS, respectively. 
        The experiments are conducted on GFNet-H-Ti.
        }
    \vspace{-0.4cm}
    \label{table: ablation all}
    \resizebox{\linewidth}{!}{
      \setlength{\tabcolsep}{4mm}{
    \begin{tabular}{l| cc | ccccc |c}
        \hline
        Methods & ETS & STS & PACS & VLCS & OfficeHome & TerraInc & DomainNet & \textbf{Avg.} \\
        \hline
        Baseline \cite{rao2021global} & * & * & 87.21 $\pm$ 0.35 & 78.90 $\pm$ 0.30 & 71.11 $\pm$ 0.18 & 50.30 $\pm$ 0.50 & 45.40 $\pm$ 0.10 & 66.59 \\
        \hline
        Both Origin & - & - & 88.79 $\pm$ 0.16 & 79.09 $\pm$ 0.47 & 71.44 $\pm$ 0.13 & 51.72 $\pm$ 0.19 & 45.49 $\pm$ 0.05 & 67.31 \\
        Edge + Origin & \checkmark & - & 89.73 $\pm$ 0.47 & 79.29 $\pm$ 0.69 & 71.82 $\pm$ 0.03 & 52.13 $\pm$ 0.70 & 45.62 $\pm$ 0.08 & 67.72 \\
        Origin + Shuffle & - & \checkmark & 89.95 $\pm$ 0.17 & 79.43 $\pm$ 0.62 & 71.90 $\pm$ 0.12 & 52.43 $\pm$ 0.35 & 45.85 $\pm$ 0.08 & 67.91 \\
        \rowcolor{mygray} \textbf{Edge + Shuffle} & \checkmark & \checkmark & \textbf{90.52 $\pm$ 0.21} & \textbf{79.71 $\pm$ 0.22} & \textbf{72.33 $\pm$ 0.13} & \textbf{52.56 $\pm$ 0.63} & \textbf{45.95 $\pm$ 0.40} & \textbf{68.21} \\
        \hline
    \end{tabular}}}
    \end{center}
    \vspace{-0.6cm}
  \end{table*}

\begin{table}[tb!] 
    \begin{center}
      \caption{
        Comparison (\%) between the ``Token Filtering'' strategy and the ``Fourier Masking'' strategy. 
        ``Token Filtering'' involves filtering out edge-irrelevant tokens from feature maps, whereas ``Fourier Masking'' refers to leveraging the high-frequency spectrum of feature maps for token mixing. The experiments are conducted on PACS using GFNet-H-Ti.
        }
    \vspace{-0.4cm}
    \label{table: Token Filtering strategy}
    \resizebox{\linewidth}{!}{
    \begin{tabular}{l|cccc|c}
        \hline
        Method & Art & Cartoon & Photo & Sketch & Avg. \\
        \hline
        Baseline \cite{rao2021global} & 89.44 & 81.34 & 97.99 & 80.06 & 87.21 \\
        Fourier Masking & 90.71 & 84.96 & 98.27 & 86.59 & 90.13 \\
        Token Filtering (SETA) & \textbf{90.95} & \textbf{85.55} & \textbf{98.37} & \textbf{87.19} & \textbf{90.52} \\
        \hline
    \end{tabular}}
    \end{center}
    \vspace{-0.4cm}
  \end{table}

\textbf{Ablation study on each component.} 
 We conduct the ablation study to investigate the efficacy of Activation-based Edge Token Selection (ETS) and Shape-disrupted Token Shuffling (STS) modules. 
 Specifically, ``\textit{Both Origin}'' is to mix representations of the original sample with randomly selected sample representations. 
 For ``\textit{Edge + Origin}'', we extract edge tokens of the original sample and mix them with a randomly selected sample without token shuffling.
 ``\textit{Origin + Shuffle}'' involves mixing the original sample representations with the token-shuffled sample representations. 
 The experiments are conducted on the DomainBed benchmark using the GFNet-H-Ti backbone.
 As depicted in Tab.~\ref{table: ablation all}, both ETS and STS consistently achieve improvements across five datasets, \textit{e.g.}, the ``Edge + Origin'' with ETS outperforms the ``Both Origin'' configuration by $0.41\%$ ($67.72\%$ vs. $67.31\%$), and the ``Origin + Shuffle'' with STS outperforms the ``Both Origin'' by $0.60\%$ ($67.91\%$ vs. $67.31\%$) on average.
 Furthermore, our SETA framework (\textit{i.e.}, ``Edge + Shuffle'') achieves the best average performance, surpassing the Baseline by $1.62\%$ ($68.21\%$ vs. $66.59\%$). 
 These results affirm that the ETS and STS modules complement and mutually enhance each other.

  \begin{table}[tb!]
    \centering
    \caption{
        \textcolor{revised_checked}{
        Effects of different weights $w$ for edge-irrelated tokens in our SETA. The experiment is conducted on the PACS dataset with GFNet-H-Ti as the backbone.}
    }
    \vspace{-0.2cm}
    \resizebox{\linewidth}{!}{
        \setlength{\tabcolsep}{3mm}{
    \begin{tabular}{c|cccc|c}
        \hline
        Weight $w$ & Art & Cartoon & Photo & Sketch & Avg. \\
        \hline
        1.00 & 90.86 & 85.26 & 98.34 & 85.35 & 89.95 \\
        0.75 & 90.16 & 84.51 & 98.28 & 86.16 & 89.78 \\
        0.50 & 90.64 & 85.11 & 98.13 & 86.19 & 90.02 \\
        0.25 & 90.48 & 85.30 & 98.28 & 86.23 & 90.07 \\
        0.00 & \textbf{90.95} & \textbf{85.55} & \textbf{98.37} & \textbf{87.19} & \textbf{90.52} \\
        \hline
    \end{tabular}
    }}
    \label{tab:non-edge weights}
    \vspace{-0.4cm}
  \end{table}

\textbf{Why filtering edge-irrelated tokens.}
\label{exp:EST strategy}
  To investigate the effect of the activation-based edge tokens selection (ETS) strategy, we here compare it with the ``Fourier Masking'' strategy that directly low-frequency filtering representation for token mixing. We here denote the ETS strategy as ``Token Filtering'', which preserves a subset of original tokens that contain edge information. 
  The experiments are conducted on the PACS dataset with the GFNet-H-Ti backbone. 
  As presented in Tab.~\ref{table: Token Filtering strategy}, the ``Token Filtering'' strategy achieves stable improvements from the ``Fourier Masking'' strategy, outperforming by $0.39\%$ ($90.52\%$ vs. $90.13\%$) on average. 
  The results demonstrate that our ETS strategy could effectively protect useful semantic information within edge-related tokens, thus helping the model learn domain-invariant features.

\textcolor{revised_checked}{
To study whether simply masking the edge token affects feature extraction, we also conduct an experiment that leverages weights to weight non-edge tokens instead of roughly masking them. 
Specifically, we introduce a hyperparameter $w$ to assign weights to non-edge tokens, where the Eq.~(\ref{Eq:edge tokens}) is re-formulated as: $z_i^e = \mathcal{M}_{edge} \odot z_i + w \cdot (1 - \mathcal{M}_{edge}) \odot z_i$. We investigate the impact of $w$ on model generalization, where $w \in \{0.00,0.25,0.50, 0.75, 1.00\}$. Notably, $w=0.00$ corresponds to the non-edge masking method (\textit{i.e.}, SETA), while $w=1.00$ is the baseline without SETA.
As presented in Tab.~\ref{tab:non-edge weights}, SETA (\textit{i.e.}, $w=0.00$) achieves the best performance among all variants, \textit{e.g.}, exceeding the variant with $w=0.75$ by $0.74\%$ ($90.52\%$ vs. $89.78\%$). 
The results indicate that SETA could adaptively retain tokens that contain most of the edge information while introducing the non-edge tokens with weight $w$ could introduce harmful domain-specific information.}

\begin{figure}[tb!]
    \centering
        \subfloat[Inserted posotions.]{
          \includegraphics[width=0.49\linewidth]{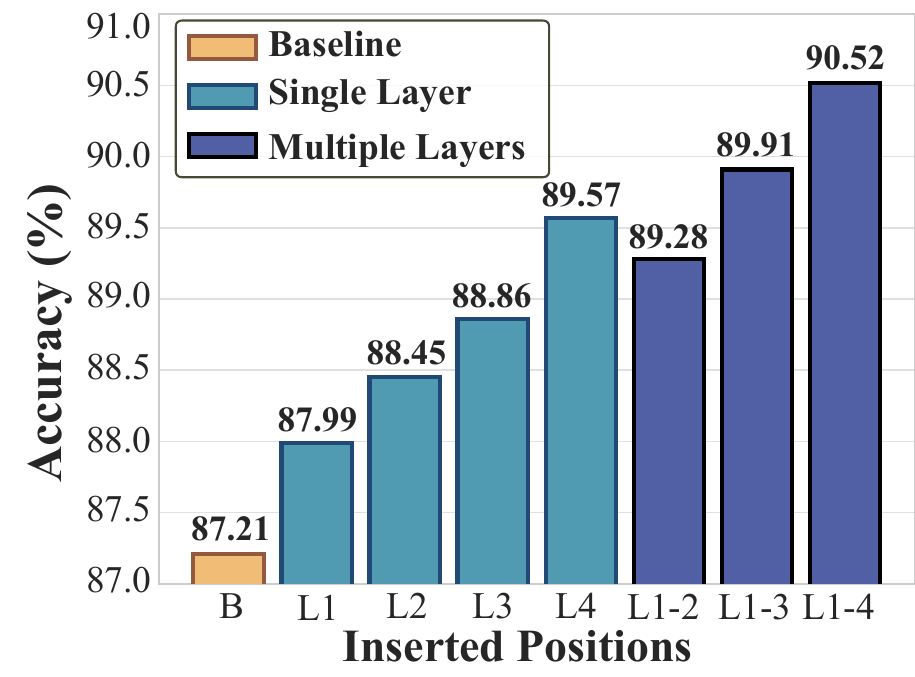}
          \label{fig:position}
        }
        \subfloat[Token selection ratio.]{
          \includegraphics[width=0.49\linewidth]{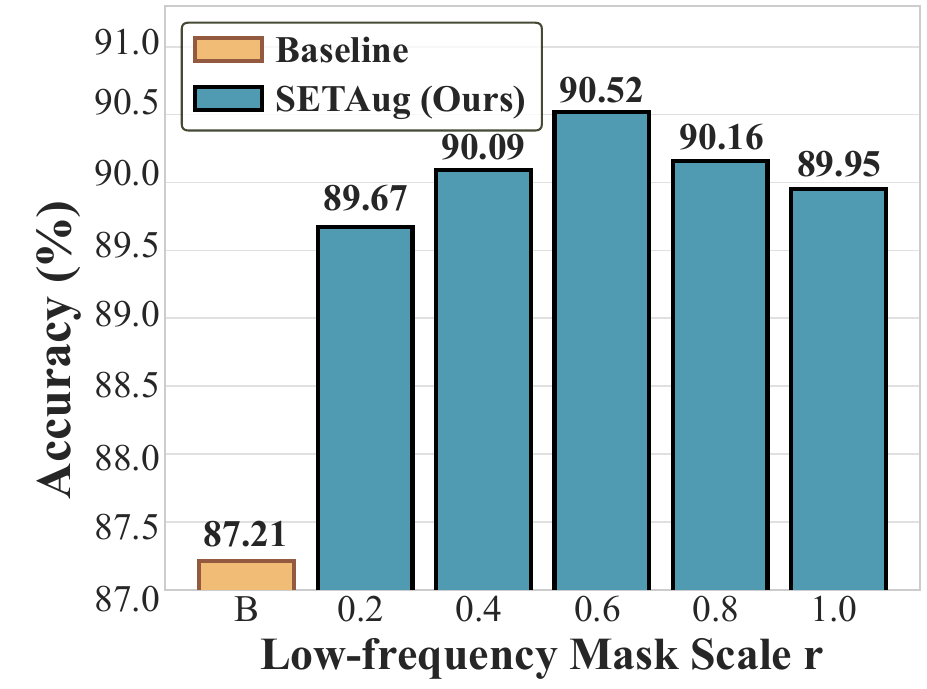}
          \label{fig:scale PACS}
        }
        \caption{
            \textcolor{revised_checked}{
        Effects of hyper-parameters including inserted positions and low-frequency mask scale $r$ in SETA.
        The experiments are conducted on PACS with GFNet-H-Ti backbone. L$1$-$4$ are four transformer layers of the network.}
        }
        \label{fig:hyper}
        \vspace{-0.5cm}
\end{figure}

\begin{figure}[tb!]
    \centering
        \subfloat[On PACS.]{
          \includegraphics[width=0.49\linewidth]{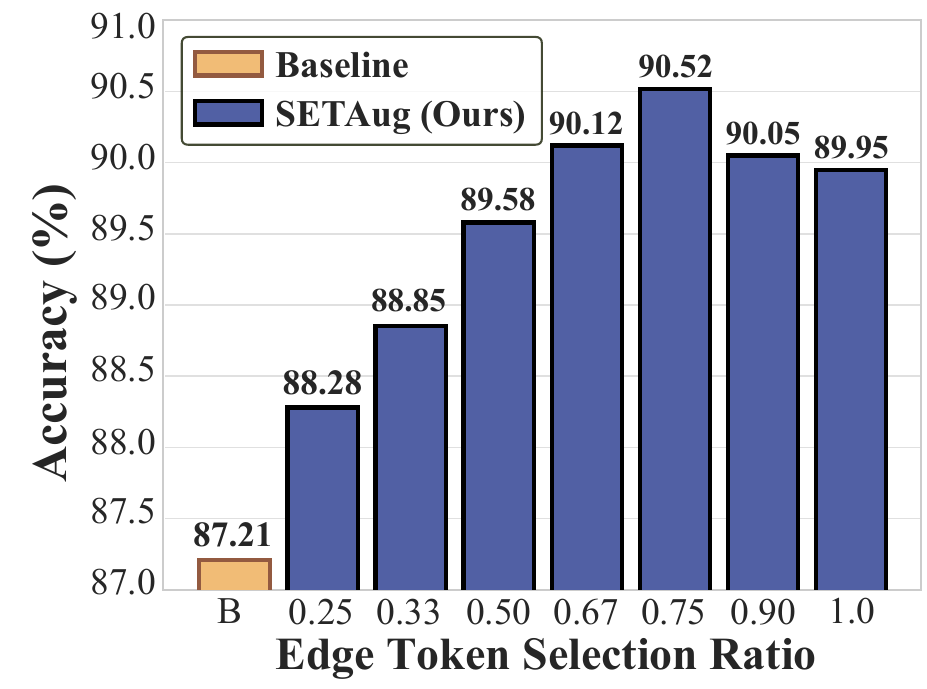}
          \label{fig:ratio PACS}
        }
        \subfloat[On OfficeHome.]{
            \includegraphics[width=0.49\linewidth]{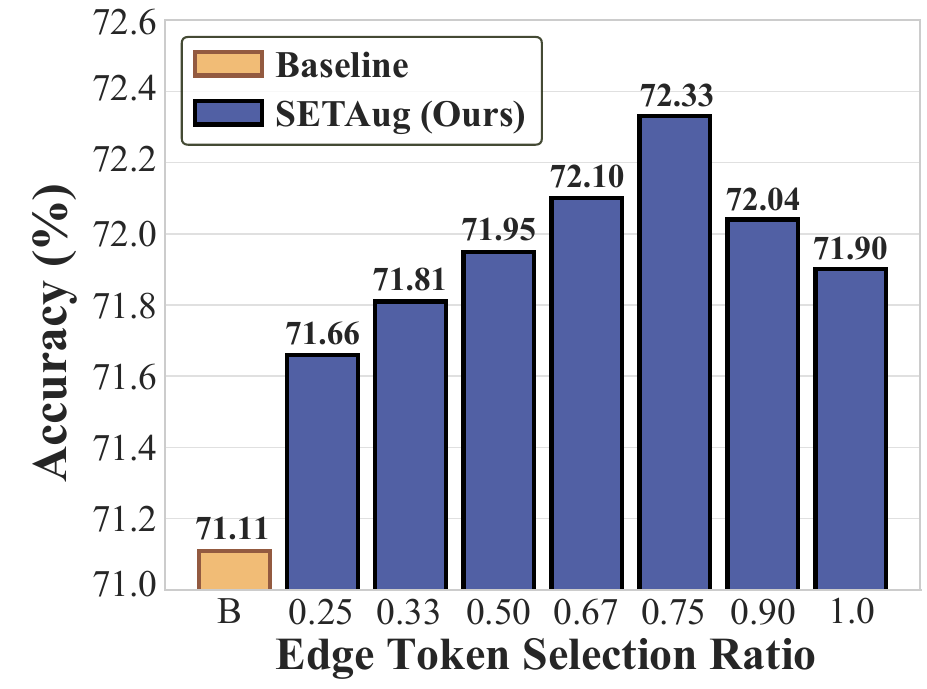}
            \label{fig:ratio OH}
          }
        \caption{
            \textcolor{revised_checked}{Effects of the hyper-parameter edge token selection ratio $p$ in SETA. The experiments are conducted on PACS and OfficeHome with GFNet-H-Ti.}
        }
        \label{fig:edge ratio}
        \vspace{-0.3cm}
\end{figure}

\textbf{Effects of SETA inserted positions.}
To assess the impact of the positions where SETA is inserted, given that a standard hierarchical model (for both GFNet-H-Ti and Swin-T) comprises four transformer layers as L$1$-L$4$, 
we evaluate the effects of applying SETA to different positions accordingly.
As shown in Fig.~\ref{fig:hyper} (a), each individually inserted SETA results in improvements compared to the baseline, with the best performance achieved when SETA is applied to all layers.
The results suggest that augmenting shape bias in all network layers can effectively boost the generalization ability of the model.
Therefore, we integrate SETA into all transformer layers for all our experiments.

\textcolor{revised_checked}{
\textbf{Sensitivity to low-frequency mask scale.} 
We analyze the effect of the hyperparameter $r$, which determines the extent to which low-frequency components are discarded, on the generalization ability of the model. 
We vary $r$ within the range $\{0.2, 0.4, 0.6, 0.8, 1.0\}$ on PACS. 
As shown in Fig.~\ref{fig:hyper} (b), the best performance is achieved with $r=0.6$ for both datasets, suggesting that this setting effectively filters out low-frequency style information while preserving high-frequency edge information. 
Thus, we use $r=0.6$ across all datasets.}

\textbf{Sensitivity to edge token selection ratio.} 
We here investigate the sensitivity of SETA to the hyperparameter $p$, which represents the percentage of edge tokens extracted from the original samples. 
\textcolor{revised_checked}{We vary $p$ within the range $\{0.25, 0.4, 0.5, 0.67, 0.75, 0.9, 1.0\}$ on both PACS and OfficeHome. As shown in Fig.~\ref{fig:edge ratio}, as $p$ increases from $0.25$ to $0.75$, the accuracy improves from $88.28\%$ to $90.52\%$ on PACS and from $71.66\%$ to $72.33\%$ on OfficeHome, consistently surpassing the baseline by a significant margin.}
These results prove the stability of our method in enhancing the shape bias of the model. However, if $p$ is too large, \textit{i.e.}, retaining too many texture tokens, the remaining shape-irrelevant features may cause the model to learn domain-specific information, leading to performance degradation. The optimal performance is achieved with $p$ set to $0.75$, which we adopt as the default setting in our experiments for all datasets.

\begin{table}[tb!]
  \centering
  \caption{Fractions (\%) of Dimensionality that encodes shape cues or texture cues \cite{islam2021shape,tripathi2023edges}.
  The texture and shape factors are estimated for the last-stage representation with the total dimension as $512$.
  The remaining dimensions are allocated to the ``residual'' factor. 
  }
  \vspace{-0.2cm}
  \resizebox{\linewidth}{!}{
  \begin{tabular}{l|cc}
      \hline
      Models & Shape Factor ($\%\uparrow$)  & Texture Factor ($\%\downarrow$)\\
      \hline
      Baseline \cite{rao2021global} & 31.23 & 19.24 \\
      \textbf{SETA (Ours)} & \textbf{33.83 ($\uparrow$ 2.60)} & \textbf{18.67 ($\downarrow$ 0.57)} \\
      \hline
      DSU \cite{li2022uncertainty} & 33.43 & 18.82 \\
      \textbf{SETA-S-DSU (Ours)} & \textbf{34.31 ($\uparrow$ 0.88)} & \textbf{18.54 ($\downarrow$ 0.28)} \\
      \hline
      ALOFT \cite{guo2023aloft} & 33.25 & 18.77 \\
      \textbf{SETA-S-ALOFT (Ours)} & \textbf{34.15 ($\uparrow$ 0.90)} & \textbf{18.56 ($\downarrow$ 0.21)} \\
      \hline
  \end{tabular}}
  \label{tab:Factors}
  \vspace{-0.3cm}
\end{table}

\textbf{Effectiveness on other architectures.} 
To further validate the effectiveness of our method, we conduct experiments on various network architectures. 
We first provide the results of our SETA on the representative ViT model, \textit{i.e.}, DeiT-S \cite{touvron2021training} with $22$M parameters. 
For implementation details, we divide the $12$ network layers of DeiT-S into four blocks by $\{2, 2, 6, 2\}$ as \cite{rao2021global,liu2021swin}, with SETA inserted into each block employing the same hyperparameters as previous experiments. 
The experiments are conducted on the PACS dataset.
As presented in Tab.~\ref{tab:domainbed-DeiT-CNN}, on the advanced DeiT-S network, SETA still significantly enhances the model performance, achieving the improvement of $3.28\%$ ($89.13\%$ vs. $85.85\%$) from the baseline. 
The results highlight the effectiveness of our SETA method to enhance the shape bias of the model. 

Moreover, while our SETA is a token-level augmentation designed for ViTs, we also design a variant for CNNs, denoted as SETA-CNN.
Since CNNs utilize local convolutions for feature extraction, lacking patch-level token learning, the SETA-CNN is essentially a pixel-level feature augmentation.
In detail, SETA-CNN treats each spatial pixel of feature maps in CNNs as a ``token'', selecting edge-related pixels based on their values in edge maps, shuffling spatial pixel positions to generate shape-disrupted noise, and mixing them to produce pixel-level augmented representations. 
SETA-CNN is inserted after all residual blocks of the network. 
The experiments are conducted on the conventional ResNet-$50$, and the results are shown in Tab.~\ref{tab:domainbed-DeiT-CNN}. 
Our SETA achieves the improvement of $1.68\%$ ($87.18\%$ vs. $85.50\%$) from the baseline, which validates the generalizability of our method.
However, CNNs primarily rely on local convolution for learning and lack exploration of spatial dependencies, while our method primarily enhances the shape bias of the model through feature extraction within tokens and spatial dependencies among tokens.
The difference in architecture may constrain the performance of SETA on CNNs. 
In conclusion, the experimental results demonstrate the generalizability and effectiveness of our method.

\renewcommand{\thefootnote}{\fnsymbol{footnote}} 
\begin{table}[tb!]
    \centering
    \caption{
        Performance (\%) of SETA on othere architectures, including ResNet$50$ and DeiT. The experiment is conducted on the PACS dataset.
    }
    \vspace{-0.2cm}
    \resizebox{\linewidth}{!}{
        \setlength{\tabcolsep}{3mm}{
    \begin{tabular}{l|cccc|c}
        \hline
        Method & Art & Cartoon & Photo & Sketch & Avg. \\
        \hline
        DeiT-S \cite{touvron2021training} & 87.55 & 82.16 & 98.45 & 75.24 & 85.85 \\
        + SETA {\scriptsize (Ours)} & \textbf{89.30} & \textbf{86.61} & \textbf{98.52} & \textbf{82.08} & \textbf{89.13} \\
        \hline
        ResNet$50$ \cite{he2016deep} & 84.70 & 80.80 & 97.20 & 79.30 & 85.50 \\
        + SETA {\scriptsize (Ours)} & \textbf{87.49} & \textbf{83.05} & \textbf{97.28} & \textbf{80.92} & \textbf{87.18} \\
      \hline
    \end{tabular}
    }}
    \label{tab:domainbed-DeiT-CNN}
    \vspace{-0.2cm}
  \end{table}

\textbf{Quantitative analysis of shape bias.} 
\label{sec:Quantitative analysis}
To quantify the amount of texture and shape information, we follow the pioneering work \cite{islam2021shape} and estimate the number of channels that represent the shape or texture concept. 
Given a representation $z \in \mathbb{R}^{N \times C}$, where $N$ is the token number and $C$ is the channel number.
We initially construct image pairs $\{(z_a, z_b)\}$ with similar semantic concept $F$ (shapes or textures), by Fourier-based transformations \cite{xu2021fourier}, \textit{i.e.}, shape-bias pairs with the same phase but different amplitudes, and texture-bias pairs with the same amplitude but different phases. 
We approximate the mutual information between $z_a$ and $z_b$ with their correlation for each channel, summing over which yields a relative score for concept $F$, formulated as below:
\vspace{-0.05cm}
\begin{equation}
    s_F = \frac{1}{C}\sum_{i=1}^C \frac{{\rm Cov}(z_{a}^{i}, z_{b}^{i})}{\sqrt{{\rm Var}(z_a^i){\rm Var}(z_b^i)}}.
\end{equation}
where $s_F$ is in $[-1, 1]$.
The remaining dimensions are allocated to residual factor $s_{r}$ with the maximum score $1$. 
For the set of concept factors $\{s_r, s_s, s_t\}$, we conduct a softmax to get the dimensionality fraction for the $F$ concept: $\rho_F = e^{s_F} / \sum_{k=1}^{3} e^{s_k}$.
As shown in Tab.~\ref{tab:Factors}, both the SOTA DG methods present a higher shape factor and lower texture factor than the baseline.
The improvements are attributed to their perturbation of styles, potentially amplifying underlying shape bias.
However, since they do not explicitly distinguish between global shape and local edges, their improvement to shape bias is still limited.
Differently, \textit{our SETA perturbs local edges to enhance model learning of global shapes, thus effectively increasing shape factor while reducing texture factor}, 
\textit{e.g.}, boosting shape factor by $0.88\%$ ($34.31\%$ vs. $33.43\%$) from DSU and $0.90\%$ ($34.15\%$ vs. $33.25\%$) from ALOFT. 
The results align well with the discussion in Sec.~\ref{Theoretical analysis}, proving that our SETA can effectively enhance the shape sensitivity, thus improving the generalization ability of the model.

\begin{table}[tb!] 
  \begin{center}
  \caption{
      The distribution gap between source and target domains. The experiments are conducted on the PACS dataset with GFNet-H-Ti as the backbone. The lower the value, the smaller the distribution gap.
      }
  \vspace{-0.1cm}
  \label{table: domain gap}
  \resizebox{\linewidth}{!}{
  \begin{tabular}{l| cccc | c}
      \hline
      Methods & Art & Cartoon & Photo & Sketch & Avg. \\
      \hline
      Baseline \cite{rao2021global} & 7.77 & 9.19 & 6.95 & 15.88 & 9.95 \\
      \rowcolor{mygray} \textbf{SETA} & \textbf{4.40} & \textbf{5.50} & \textbf{5.02} & \textbf{7.15} & \textbf{5.52} \\
      \hline
      DSU \cite{li2022uncertainty} & 5.95 & 7.03 & 4.84 & 8.98 & 6.70 \\
      \rowcolor{mygray} \textbf{SETA-S-DSU} & \textbf{4.79} & \textbf{5.48} & \textbf{4.23} & \textbf{7.03} & \textbf{5.38} \\
      \hline
      ALOFT \cite{guo2023aloft} & 5.36 & 7.11 & 5.27 & 8.38 & 6.53 \\
      \rowcolor{mygray} \textbf{SETA-S-ALOFT}  & \textbf{4.84} & \textbf{5.25} & \textbf{4.18} & \textbf{6.61} & \textbf{5.22} \\
      \hline
  \end{tabular}}
  \end{center}
  \vspace{-0.3cm}
\end{table}

\begin{figure*}[tb!]
    \centering
    \includegraphics[scale=0.53]{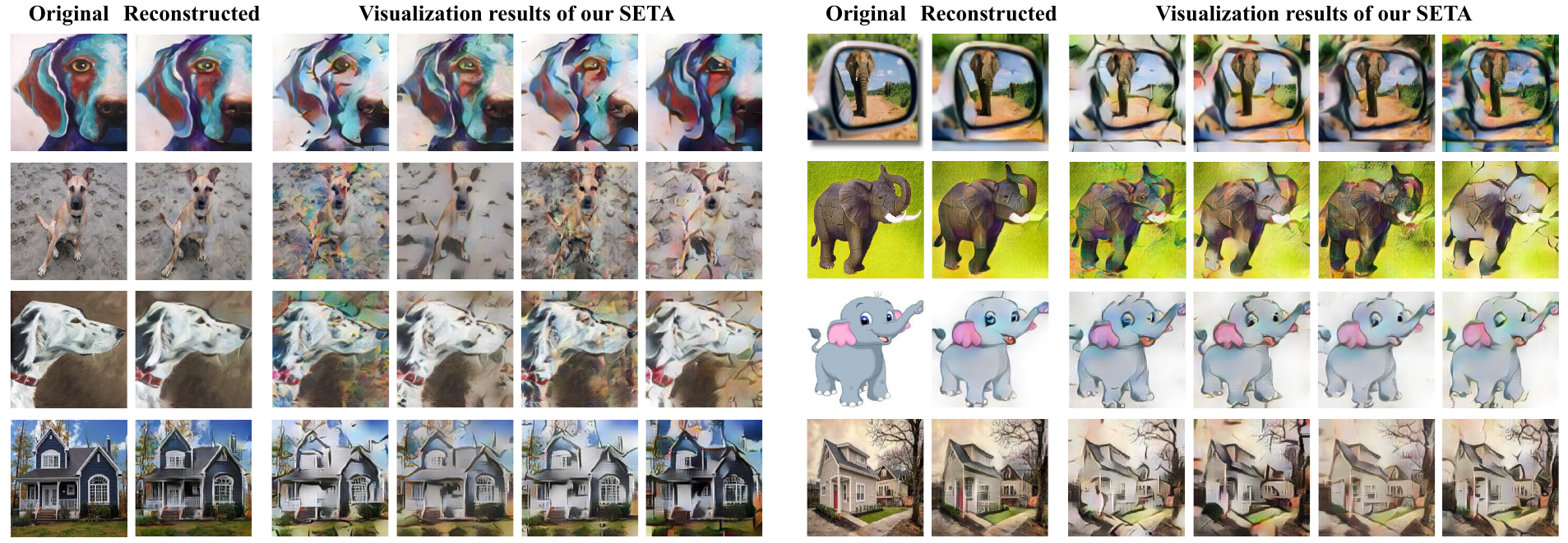}
    \vspace{-0.3cm}
        \caption{
            The visualization results on diverse synthetic changes produced by our SETA method. We follow \cite{huang2017arbitrary} and use the first few layers of a fixed VGG-$19$ to encode the images. The SETA is used to perform augmentation in the feature space. The representations are inverted to synthetic images by a decoder.
        }
        \label{fig:Visualization}
        \vspace{-0.2cm}
\end{figure*}

\textbf{Domain gap between source and target domains.} 
\label{sec:domain gap}
To investigate the impact of SETA on the domain gap between source and target domains, we estimate the domain distance $d_{H\Delta H}(D_s^{aug}, D_t)$ in Eq.~(\ref{eq:aug generalization 1}) by evaluating the domain discrepancy of the feature maps extracted by the last layer of the model. 
Let $\overline{f_s^k}$ denote the averaged feature maps of all samples from the $k$-th source domain, and $\overline{f_t}$ represents the averaged feature maps of all samples from the target domain. 
The domain discrepancy between the source and target domains is then computed as: $d = \frac{1}{K} \sum_{k=1}^{K} \| \overline{f_s^k} - \overline{f_t}\|_2.$
The experiments are conducted on the PACS dataset using the GFNet-H-Ti architecture, in which we select a domain as the target domain and evaluate the distance between it and the other source domains. 
As shown in Tab.~\ref{table: domain gap}, both DSU and ALOFT demonstrate the ability to reduce the domain gap between source and target domains, indicating that perturbing style features could enhance the model robustness to domain shifts.
However, these methods overlook the fact that edge features could also encode domain-specific information (\textit{e.g.}, shape-irrelevant local edges), leading to an insufficient narrowing of the domain gap. 
In contrast, our proposed method explicitly encourages the model to learn the holistic shape of the object by perturbing local edges. 
Consequently, our method effectively reduces the domain gap between source and target domains, \textit{e.g.}, achieving a reduction of $4.43$ ($5.52$ vs. $9.95$) compared to the Baseline. 
Notably, with the target domain being Sketch emphasizing shape features, our SETA achieves a significant reduction in domain distance between Sketch and the other three source domains, reducing $8.73$ ($7.15$ vs. $15.88$) from the Baseline.
The result indicates that our method could effectively enhance the model learning of shape representation.
Combined with existing augmentation methods, our SETA further reduces domain gaps, thereby improving the model generalization.
The results empirically validate the theoretical conclusions outlined in Sec.~\ref{sec:theorem_and_proposition}.

\begin{table}[tb!] 
    \begin{center}
    \caption{
        \textcolor{revised_checked}{
        The computational overhead of SETA and its two variants. The experiments are conducted on the PACS dataset.}
        }
    \vspace{-0.4cm}
    \label{table: computational overhead}
    \resizebox{\linewidth}{!}{
        \setlength{\tabcolsep}{1mm}{
    \begin{tabular}{l| ccc}
        \hline
        Methods & Training Time (s) & GPU Mem (GB) & Performance (\%) \\
        \hline
        \multicolumn{4}{c}{Swin-T} \\
        \hline
        Baseline \cite{liu2021swin} & 0.220 & 9.827 & 66.92 \\
        \rowcolor{mygray} \textbf{SETA} & 0.226 & 10.328 & 67.50 ($\uparrow 0.58$) \\
        \hline
        DSU \cite{li2022uncertainty} & 0.235 & 11.207 & 67.29 \\
        \rowcolor{mygray} \textbf{SETA-S-DSU} & 0.236 & 11.829 & 67.81 ($\uparrow 0.52$) \\
        \hline 
        ALOFT \cite{guo2023aloft} & 0.242 & 12.026 & 67.38 \\
        \rowcolor{mygray} \textbf{SETA-S-ALOFT} & 0.246 & 12.537 & 68.00 \textbf{($\uparrow 0.62$)} \\
        \hline 
        \hline 
        \multicolumn{4}{c}{GFNet-H-Ti} \\
        \hline 
        Baseline \cite{rao2021global} & 0.153 & 8.009 & 66.59 \\
        \rowcolor{mygray} \textbf{SETA} & 0.154 & 8.227 & 68.21 \textbf{($\uparrow 1.62$)} \\
        \hline
        DSU \cite{li2022uncertainty} & 0.181 & 8.245 & 67.33 \\
        \rowcolor{mygray} \textbf{SETA-S-DSU} & 0.184 & 8.467 & 68.36 \textbf{($\uparrow 1.03$)} \\
        \hline
        ALOFT \cite{guo2023aloft} & 0.197 & 8.366 & 68.04 \\
        \rowcolor{mygray} \textbf{SETA-S-ALOFT} & 0.200 & 8.947 & 68.58 \textbf{($\uparrow 0.54$)} \\
        \hline
    \end{tabular}}}
    \end{center}
    \vspace{-0.4cm}
\end{table}

\textbf{Computational overhead of SETA.}
\textcolor{revised_checked}{
To showcase the computational efficiency of SETA, we conduct a comparative analysis including training time per iteration and GPU memory requirements. 
We compare our methods with the baseline \cite{touvron2021training} and SOTA augmentation methods in DG, \textit{i.e.}, DSU \cite{li2022uncertainty} and ALOFT \cite{guo2023aloft}.
The results are shown in Tab.~\ref{table: computational overhead}. 
Specifically, for SETA, we observe a marginal increase in both training time and GPU memory requirement, \textit{e.g.}, on Swin-T architecture, there is an incremental time of $0.006$ seconds per iteration ($0.226$ seconds vs. $0.220$ seconds), accompanied by an additional memory requirement of $0.040$ GB ($9.867$ GB vs. $9.827$ GB) during training.
Moreover, when SETA is synergistically employed with existing DG augmentation methods, our variants still exhibit a negligible computational overhead, proving the efficiency of SETA and its two variants.}

\textbf{Visualization on the synthetic changes.}
We here provide intuitive visualizations showcasing the diverse changes generated by our SETA method. 
Following \cite{li2022uncertainty,huang2017arbitrary}, we employ a pre-defined autoencoder, \textit{i.e.}, the VGG-$19$ network, to visualize the reconstructed results. 
Specifically, SETA is incorporated into the encoder, and the augmented representations are then inverted into synthetic images by the decoder. 
The visualization results are illustrated in Fig.~\ref{fig:Visualization}.
Comparing the reconstructed images obtained with and without SETA, it is evident that our method introduces diverse variations by perturbing both the style information and local edges while preserving the shape structures. Consequently, with the integration of SETA, shape-irrelevant features are perturbed, compelling the model to learn domain-invariant global information and incentivizing the shape bias of the model.

\begin{figure}[tb!]
    \centering
    \includegraphics[scale=0.44]{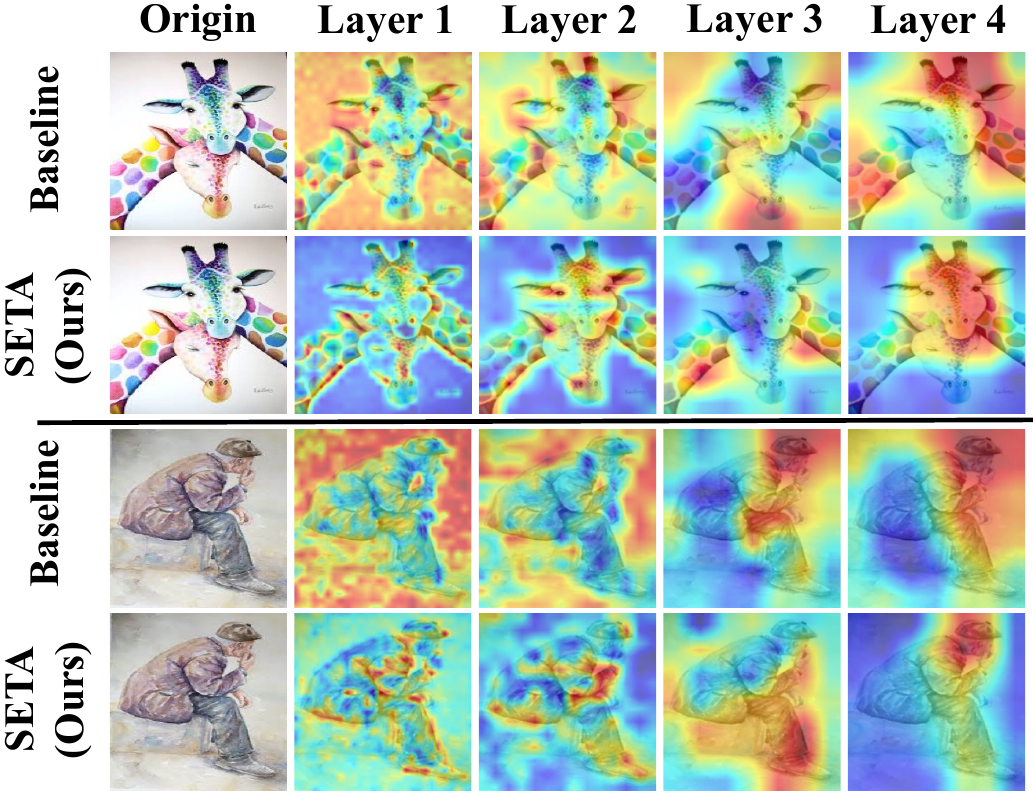}
    \vspace{-0.3cm}
        \caption{
            Visualization of attention maps in each layer of the network on the PACS dataset with Art Painting as the target domain. The backbone used in the experiment is the GFNet-H-Ti architecture. 
        }
    \vspace{-0.4cm}
        \label{fig:gradcam}
\end{figure}

\textbf{Visualization of attention maps.}
To visually verify the claim that the SETA can enhance the model's learning of global shape information, we present the attention maps of each network layer for baseline and SETA methods, employing the GradCAM technique \cite{selvaraju2017grad}. 
As presented in Fig.~\ref{fig:gradcam}, the baseline model predominantly directs its attention towards numerous shape-irrelevant regions across all network layers, which mainly encode domain-related information, thereby leading the model to overfit the source domains.
In contrast, the SETA effectively perturbs the shape-irrelevant features at each layer, compelling the model to learn a more comprehensive understanding of global shape representations. 
Taking the giraffe for example, distinct features such as its face and neck, which are pivotal for classification, are accurately captured by the SETA. 
Conversely, the baseline model tends to focus on background and texture features, thus leading to misclassification. 
These results suggest that the SETA can effectively incentivize the shape bias of the model and facilitate the learning of generalized structural information.

\section{Conclusion}
In the paper, we propose a token-level augmentation method SETA for DG to induce shape bias of the models. Different from prior CNN-based works that focus on diversifying style information within the channel dimension, our method involves discerning edge tokens and perturbing local edge features within the spatial dimension. This strategy compels the model to exhibit increased sensitivity to shape cues and improves its generalizability. To further boost model robustness to domain shifts, we extend the SETA to two advanced stylized variants with SOTA augmentation methods. Comprehensive experiments demonstrate the effectiveness of our method across various public datasets on different ViT and MLP architectures. 
\textcolor{revised_checked}{
    However, since SETA uses a Fourier-based low-frequency filtering method, its effectiveness for edge extraction might be influenced by the complexity of image textures, which could affect the distinction between class-relevant edges and noise. 
    To address this, exploring adaptive methods for feature division and extraction is a promising direction to enhance the model's shape bias further. 
    In future work, we will explore advanced methods that can learn to adaptively refine shape-irrelated edge features from complex textures to enhance generalization. 
    SETA could also be seamlessly integrated into image encoders of ViT-based vision-language models, which helps the network focus on domain-invariant shape semantics. 
    We will also further improve our SETA by leveraging text 
    information to align visual and textual representations 
    and help the model better learn generalizable semantic information.
}

\bibliographystyle{IEEEtran}
\bibliography{IEEEfull}



\begin{IEEEbiography}[{\includegraphics[width=1in,height=1.25in,clip,keepaspectratio]{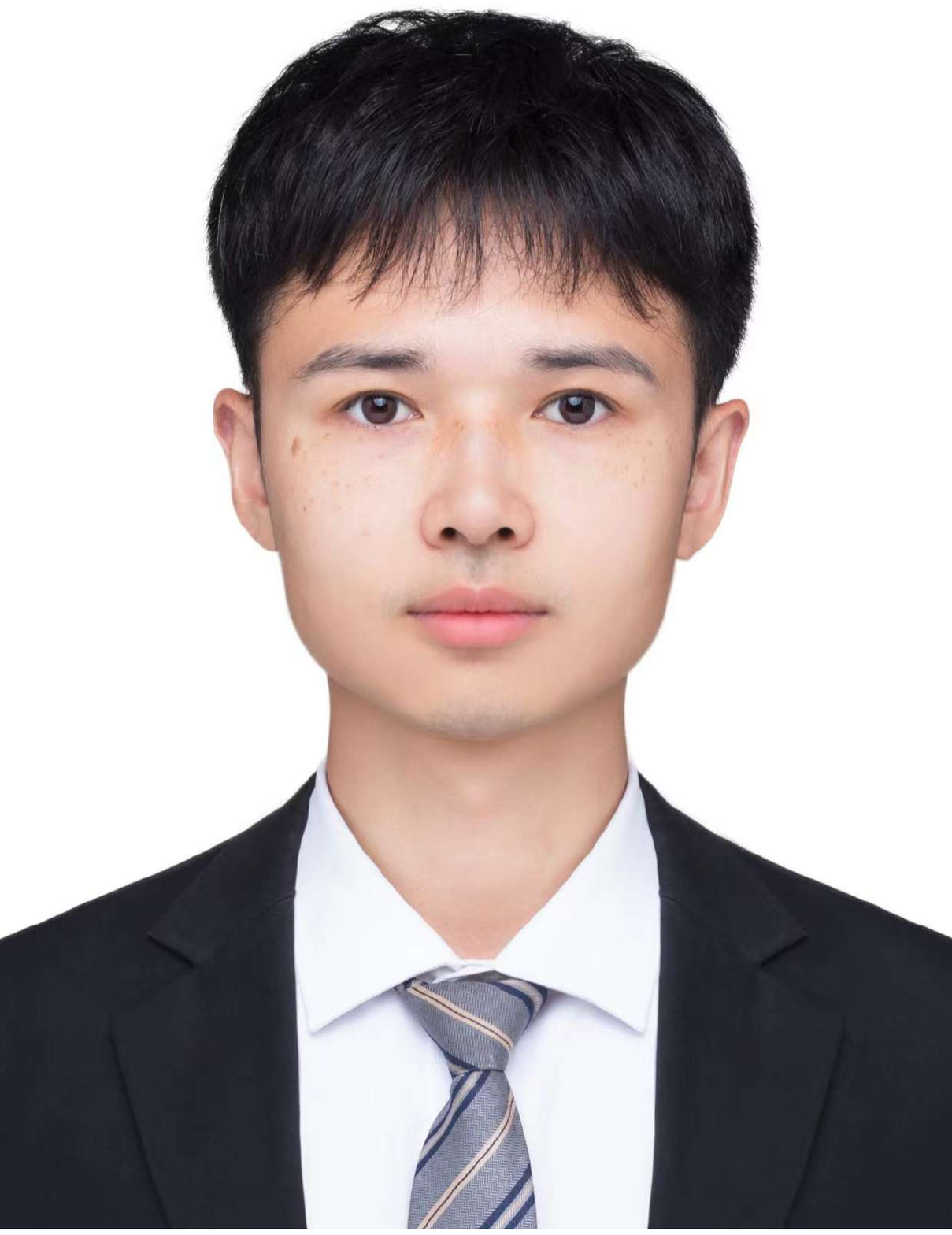}}]{Jintao Guo} is working toward the Ph.D. degree at the Department of Computer Science and Technology, Nanjing University, China. His research interests include machine learning, pattern recognition, and domain generalization.
\end{IEEEbiography}
\begin{IEEEbiography}[{\includegraphics[width=1in,height=1.25in,clip,keepaspectratio]{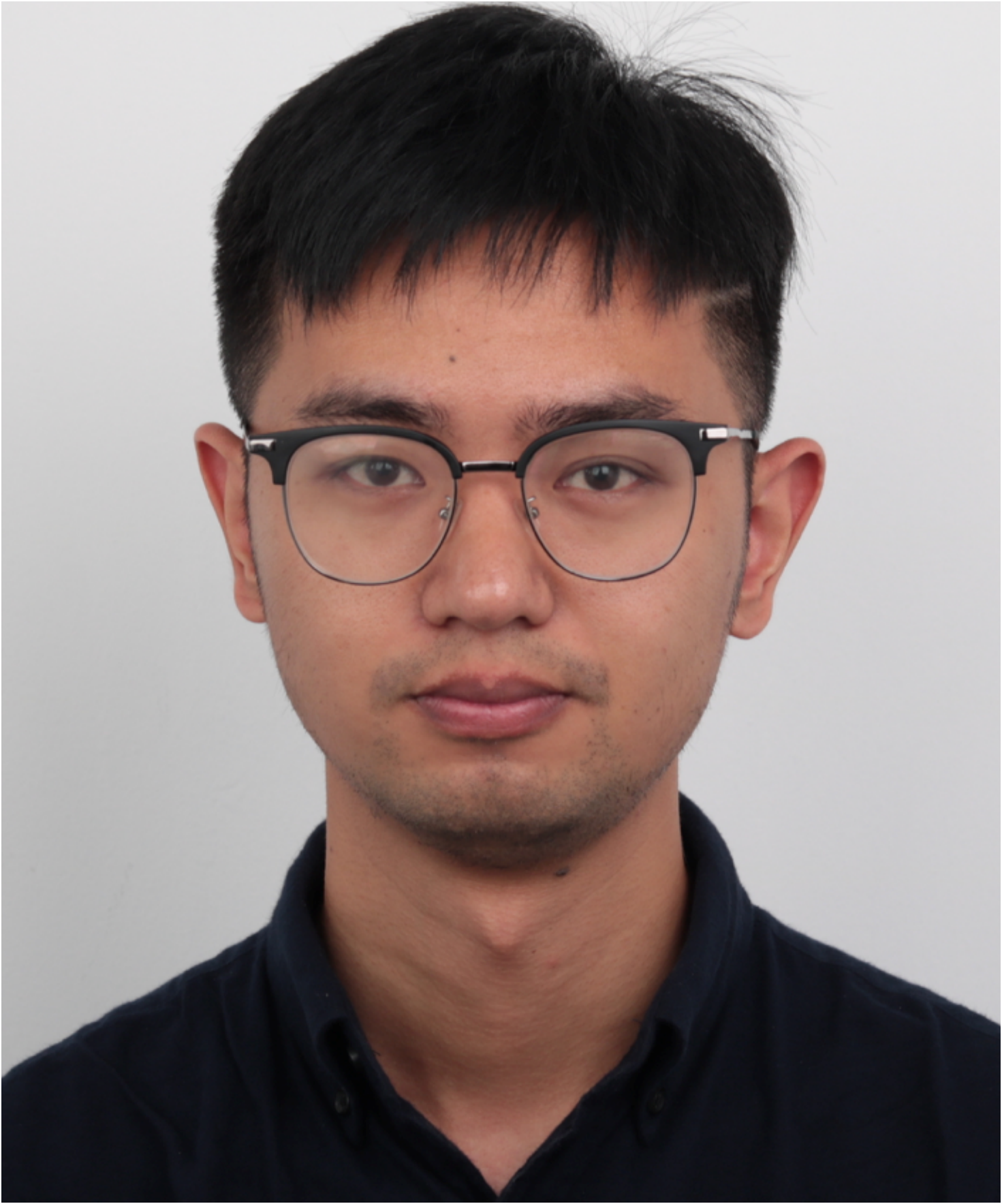}}]{Lei Qi} is currently an Associate Professor with the School of Computer Science and Engineering, Southeast University, China. His current research interests include some ML methods, such as domain adaptation, semi-supervised learning, unsupervised learning, and meta-learning. For applications, he mainly focuses on person re-identification and image segmentation.
\end{IEEEbiography}
\vfill
\begin{IEEEbiography}[{\includegraphics[width=1in,height=1.25in,clip,keepaspectratio]{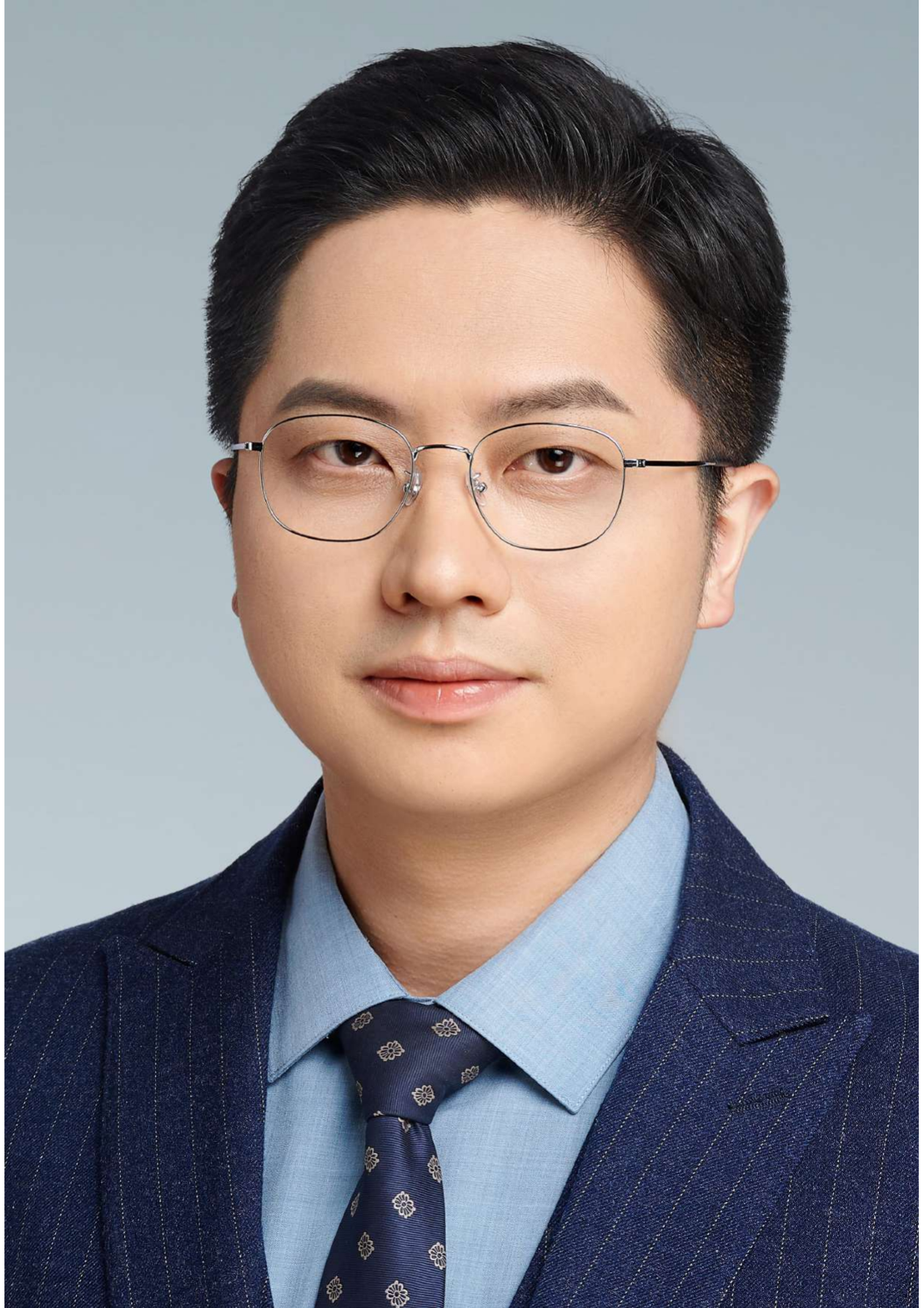}}]{Yinghuan Shi} is currently an Associate Professor at the Department of Computer Science and Technology, Nanjing University, and he is also affiliated with National Institute of Healthcare Data Science, Nanjing University. He received the B.Sc. and Ph.D. degrees both from Computer Science, Nanjing University, in 2007 and 2013, respectively. His research interests include machine learning, pattern recognition, and medical image analysis. He has published more than 60 papers in CCF-A Conference and IEEE Transactions.
\end{IEEEbiography}
\begin{IEEEbiography}[{\includegraphics[width=0.95in,height=1.25in,clip,keepaspectratio]{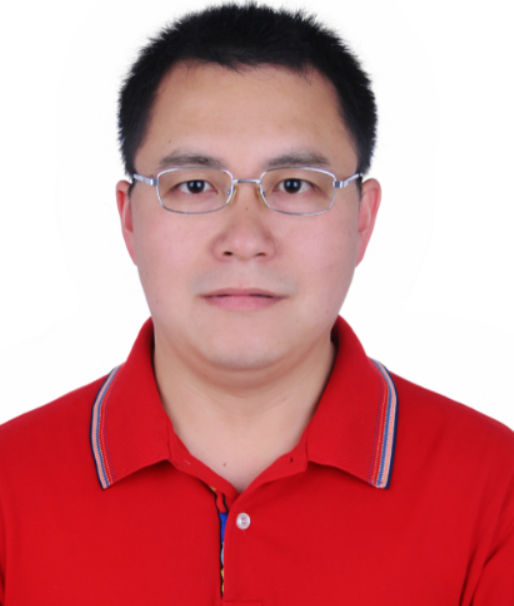}}]{Yang Gao} (Senior Member, IEEE) is a Professor in the Department of Computer Science and Technology, Nanjing University. He is currently directing the Reasoning and Learning Research Group in Nanjing University. He has published more than 100 papers in top-tired conferences and journals. He also serves as Program Chair and Area Chair for many international conferences. His current research interests include artificial intelligence and machine learning.
\end{IEEEbiography}

\vfill

\end{document}